\documentclass[11pt,a4paper]{article}% 确保 xcolor 带 table 选项，使 \rowcolor/\cellcolor 可用
\PassOptionsToPackage{table}{xcolor}

\usepackage{polyu-vclab}

\usepackage{lipsum}                % placeholder text, remove for real paper
\usepackage{amsmath}
\usepackage{booktabs}
\usepackage{multirow}
\usepackage{wrapfig}
\usepackage{makecell}     % 表格内换行 \makecell
\usepackage{xcolor}       % 颜色（lightpurple 等）
\usepackage[numbers,sort&compress]{natbib}
\definecolor{lightpurple}{RGB}{230, 220, 235}
\definecolor{lightgreen}{RGB}{220, 230, 220}
\definecolor{lightpink}{RGB}{250, 240, 240}

\setEyebrow{PolyU VCLab\,\textbullet\,Preprint 2026}
\usepackage[table]{xcolor}
\definecolor{lotuspink}{RGB}{240, 213, 218}

\setReportTag{Visual Computing Lab\,\textperiodcentered\,The Hong Kong Polytechnic University}

\papertitle{\centering PixRestore: Unified Image Restoration via \\ Pixel Diffusion Transformer}

\paperauthors{\centering%
  Lingchen Sun\equalmark\affilmark{1,2}\quad
  Rongyuan Wu\equalmark\affilmark{1,2}\quad
  Xiangtao Kong \affilmark{1,2}\quad
  Jixin Zhao \affilmark{2}\quad
  Qiaosi Yi \affilmark{1,2}\quad \\
  Yujing Sun \affilmark{1,2}\quad
  Shuaizheng Liu \affilmark{1,2}\quad
  Zhengqiang Zhang \affilmark{1,2}\quad
  Lei Zhang\correspondmark\affilmark{1,2}%
}

\paperaffil{\centering%
  \affilmark{1}\,The Hong Kong Polytechnic University\quad
  \affilmark{2}\,OPPO Research Institute
}

\papernotes{\centering%
  \equalmark\,Equal contribution.\quad
  \correspondmark\,Corresponding author (\href{mailto:cslzhang@comp.polyu.edu.hk}{cslzhang@comp.polyu.edu.hk}).%
}

\paperbadges{\centering
  \vclabbadgesolid{Project Page}{https://csslc.github.io/pixrestore-page/}\;%
  \vclabbadgesolid{Code}{https://github.com/csslc/PixRestore}\;%
}

\begin{document}
%==========================================================================

\maketitleVCLab%%% Teaser / intro figure before abstract
\begin{figure*}[htbp]
  \centering
  \includegraphics[width=1.0\textwidth]{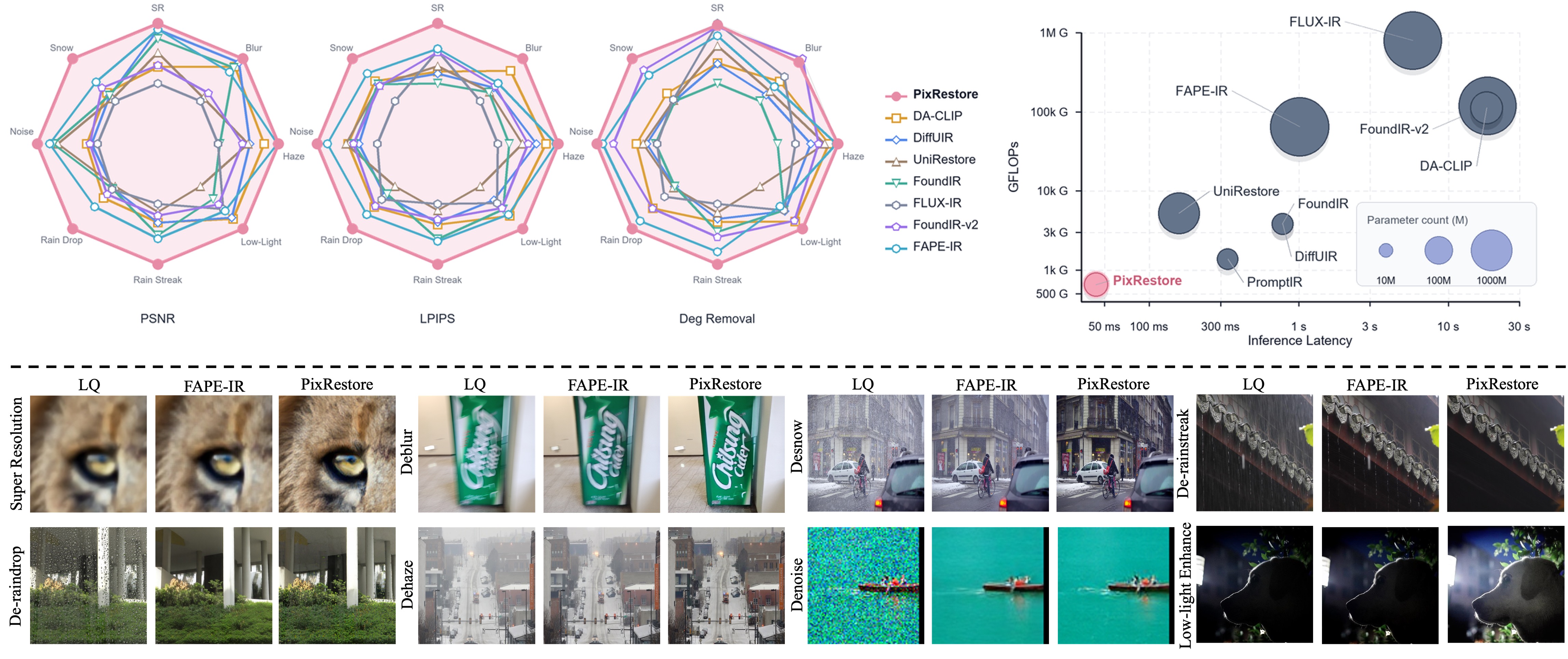}
  % \vspace{-3mm}
  \caption{\textbf{PixRestore achieves the best overall performance in terms of restoration quality, model size, and inference speed.} \textbf{Top-left:} radar charts comparing PSNR, LPIPS, and degradation removal performance across eight degradation types. \textbf{Top-right:} GFLOPs versus inference latency, where the bubble size denotes the number of parameters. \textbf{Bottom:} visual comparisons on eight restoration tasks. With only about 50M parameters and single-step inference, PixRestore achieves the best overall restoration quality while being the most efficient among diffusion-based methods.}
  \label{fig:intro}
  % \vspace{-4mm}
\end{figure*}

%-----------------------------------------------------------
% Abstract -- bordered box with red left bar (VCLab mission style)
%-----------------------------------------------------------
\begin{vclabAbstract}
\noindent\textbf{Abstract.}\;
Unified image restoration (UIR) aims to recover high-quality (HQ) content from low-quality (LQ) images with different degradations using a single model. Most recent methods adapt large pretrained text-to-image (T2I) latent diffusion models for their strong capacity and generative priors. However, the variational autoencoder (VAE) in latent T2I models may discard restoration-sensitive details, while the open-ended synthesis prior can introduce content-inconsistent artifacts. We present \textbf{PixRestore}, a VAE-free pixel-space Diffusion Transformer (DiT) for UIR, where the diffusion backbone is trained entirely from scratch, without relying on T2I pretraining.
PixRestore performs flow matching directly on patchified pixels, preserving fine-grained details while keeping the token sequence tractable.
To adapt to different degradations, PixRestore learns to predict the reliability of layer features using LQ--HQ DINO feature similarity. Features from more reliable layers are fused as dense conditioning, while less reliable layers receive stronger HQ-feature supervision to encourage degradation removal.
We train PixRestore on a large-scale corpus of diverse scenes and degradations, and further finetune it into a one-step generator using DINO-based adversarial objectives for efficient inference.
Experiments on public benchmarks and real-world test sets show that, with only about 50M parameters and single-step inference, PixRestore achieves the best overall fidelity, perceptual quality, and robustness to degradations among competing UIR models while being far more efficient. Larger PixRestore variants can further boost performance, demonstrating the scalability of our pixel-space design. Code and the curated benchmark can be found at https://github.com/csslc/PixRestore.
\end{vclabAbstract}

\keywords{Image Restoration, Pixel Diffusion, Degradation-Aware, DINO}

%==========================================================================
\section{Introduction}
%==========================================================================
Image restoration (IR) \cite{zhang2017beyond, RESIDE, uhdblurhaze} aims to recover a high-quality (HQ) image from its low-quality (LQ) counterpart corrupted by diverse and often co-occurring degradations such as noise, blur, rain, haze, and low light. Rather than training specialist models per degradation, many efforts have been devoted to pursuing unified image restoration (UIR), \textit{i.e.}, using a single model to handle a broad spectrum of image degradations \cite{UIRsurvey}. Built on CNN- and Transformer-based backbones, conventional regression-based methods~\cite{AirNet,PromptIR,MoCE-IR,DA-CLIP,li2025foundir} are efficient and have made encouraging progress. However, their deterministic $L_1$/$L_2$ objectives and limited capacity tend to yield over-smoothed results and unremoved degradations.

Diffusion models have recently been applied to UIR~\cite{diffuir,DA-CLIP} to improve perceptual quality by leveraging their strong generative capacity. Most recent works finetune pretrained text-to-image (T2I) latent diffusion models, \textit{e.g.}, FoundIR-v2 \cite{foundirv2} adapts SDXL \cite{sdxl} with MoE routing and an MLLM \cite{LLaVA} captioner, and FLUX-IR \cite{fluxir} finetunes FLUX \cite{FLUX} with reinforced ODE trajectories and cost-aware distillation. T2I pretraining brings rich priors and perceptual realism, but at three costs: (1) \textit{Lossy latent bottleneck}, as the VAE may discard image textures and details that restoration aims to preserve; (2) \textit{Objective mismatch}, as T2I priors may synthesize visually plausible but inconsistent details with the input; and (3) \textit{Redundant computation}, caused by the billion-scale backbones, MLLM planners, MoE routing, VAE coding, and iterative sampling. This motivates us to rethink the suitability of T2I models for UIR. Unlike T2I generation, which synthesizes an image from textual cues, UIR starts from LQ images, which contain rich visual cues. Therefore, UIR requires less open-ended generative capacity than T2I models, but it demands robustness to different degradations and faithful reconstruction of pixel-aligned details.

Motivated by the above observations, we present \textbf{PixRestore}, a VAE-free pixel diffusion transformer (DiT) \cite{dit, jit} for UIR. The diffusion backbone is trained from scratch without T2I pretraining. By applying flow matching directly to patchified pixels, PixRestore preserves pixel-aligned evidence while keeping the token sequence tractable. Our method removes the autoencoder overhead and substantially improves content fidelity while maintaining strong perceptual quality.
Unlike task-specific IR, UIR must handle diverse and compounded degradations, making a static global degradation representation insufficient. We therefore exploit hierarchical features from the self-supervised visual foundation model DINO \cite{dinov2} to provide adaptive guidance for restoration.
Our analysis reveals two properties of DINO features: the layers carry complementary cues, from shallow structures to deep semantics, and their reliability varies across degradation types. Accordingly, we train an adaptive layer router to predict per-layer weights from the LQ input, using LQ--HQ DINO feature similarity as supervision during training. %Higher similarity indicates that a layer is more reliable for the given input. 
The predicted weights fuse features from more reliable layers into dense conditioning, while less reliable layers receive stronger supervision from the corresponding HQ features.%, so that training focuses more on heavily degraded content.

Finally, to speed up PixRestore at inference time, we first train a multi-step PixRestore model on a large-scale corpus of diverse scenes and degradations, then finetune it into a single-step generator with DINO-based adversarial objectives. Fig.~\ref{fig:intro} compares PixRestore against existing diffusion-based UIR methods in terms of restoration quality, model size, and inference latency. With only about 50M parameters and single-step inference, PixRestore attains the best quality while being the fastest  (about 44\,ms) and most compact model among the diffusion-based methods. In addition, our experiments show that larger PixRestore variants can further improve restoration quality, confirming the scalability of our design.

Our contributions are summarized as follows:

\begin{itemize}
    \item We propose \textbf{PixRestore}, a pixel-space DiT for UIR. Working on patchified pixels, PixRestore is free of the VAE and T2I priors, improving image fidelity, perceptual quality, and model efficiency.
    \item We introduce adaptive hierarchical visual guidance, providing dense conditioning to guide restoration and supervision to stabilize training under various degradations.
    \item We finetune the multi-step model into a single-step generator via DINO-based adversarial objectives, achieving efficient inference with little quality loss.
    \item Extensive experiments show that PixRestore achieves superior fidelity, perceptual quality, and robustness on public benchmarks and real-world test sets.
\end{itemize}

\section{Related Work}
%==========================================================================

\noindent\textbf{Regression-based Unified Restoration.}
Conventional IR methods are typically developed for a specific degradation, such as noise \cite{zhang2017beyond}, blur \cite{uhdblurhaze}, rain \cite{rain1200}, haze \cite{RESIDE}, etc. UIR instead seeks a single model for multiple degradations. Existing UIR methods improve degradation adaptivity with learned degradation representations~\cite{AirNet}, prompts~\cite{PromptIR}, or expert routing~\cite{lin2024unirestorer}, etc. Methods such as PromptIR~\cite{PromptIR}, AirNet~\cite{AirNet}, and their successors show that degradation-aware modulation can substantially improve multi-task compatibility. However, these models are usually optimized as deterministic LQ-to-HQ regressors with $L_1$/$L_2$ losses, which favor conditional averages, often suppressing high-frequency details and limiting perceptual realism. Their task-level prompts or routing decisions may also generalize poorly to more complex real-world degradations. We instead model restoration as a conditional pixel-space flow and derive dense, per-image guidance from hierarchical DINO features.

\noindent\textbf{Generative Unified Restoration.}
Generative UIR methods synthesize HQ images conditioned on LQ inputs. One line of research learns the conditional generative process from scratch within the restoration task, keeping the model restoration-native~\cite{drdd, diffuir}. For example, DiffUIR~\cite{diffuir} and DA-CLIP~\cite{DA-CLIP} design restoration-specific diffusion pipelines or degradation-aware conditioning to improve fidelity across different degradations. Another line adapts pretrained T2I latent diffusion models with degradation predictors~\cite{mperceiver}, multimodal prompts~\cite{fapeir}, or routing mechanisms~\cite{foundirv2}. These methods benefit from strong generative priors, but suffer from the conflict between open-ended image synthesis and faithful restoration. In addition, the VAE compresses the input image before diffusion, removing restoration-sensitive details such as small structures, text strokes, and sharp edges. Large T2I backbones and auxiliary planners or expert modules further increase computational cost. In contrast, our proposed PixRestore operates in a patchified pixel space and uses a scalable transformer with layer-adaptive visual conditioning for efficient and faithful UIR.

\noindent\textbf{Pixel Generative Modeling.}
Diffusion is originally formulated in pixel space, whereas latent diffusion becomes dominant for the reduced generation costs~\cite{sd2.1}. Recent work has revisited VAE-free generation using improved DiT architectures \cite{jit} and loss functions \cite{ma2026pixelgen}. For UIR, the LQ image provides dense spatial correspondence to the desired output, but compressing it using a VAE can destroy useful information for restoration. Pixel-space modeling preserves that evidence, but introduces computational challenges due to the long spatial sequence. We address this trade-off by using patchification. Different from previous pixel generators developed for text-conditioned synthesis, PixRestore combines pixel-space flow modeling with degradation-aware hierarchical DINO guidance for unified restoration.

\begin{figure}[t]
  \centering
  \includegraphics[width=0.98\linewidth]{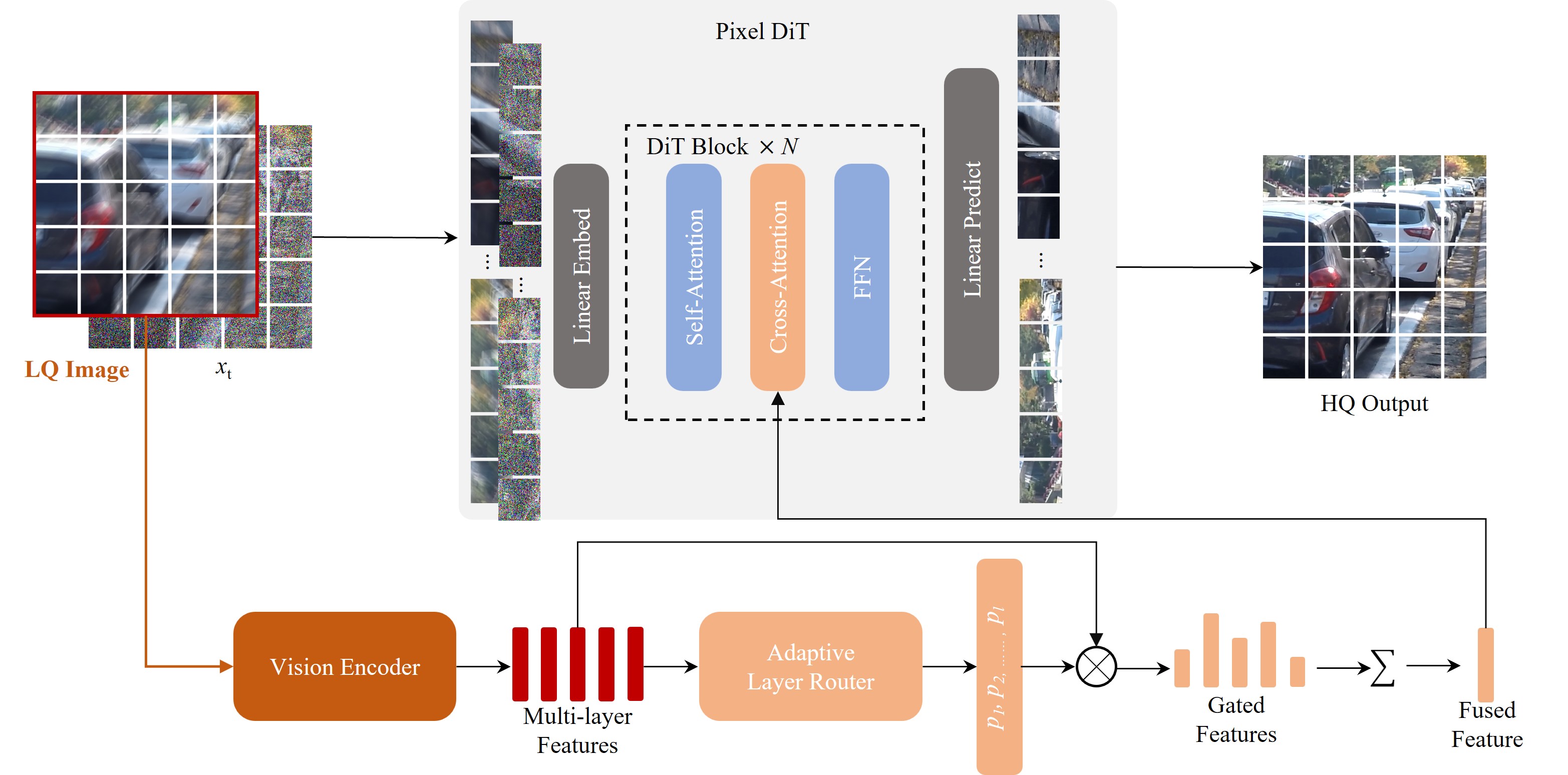}
  % \vspace{-4mm}
  \caption{Overview of PixRestore. A frozen vision encoder extracts multi-layer features from the LQ image, and an adaptive layer router predicts per-layer weights to fuse them into a single conditioning feature. The LQ image and the noisy state $x_t$ are patchified, then processed by $N$ DiT blocks, and finally decoded into the HQ output.}
  \label{fig:framework}
  % \vspace{-5mm}
\end{figure}

%==========================================================================
\section{Method}
Let \(y_{hq} \in [-1,1]^{3\times H\times W}\) be an HQ image and \(y_{lq}\) its LQ counterpart corrupted by degradations such as noise, blur, haze, rain, or low light. UIR aims to learn a single model that maps \(y_{lq}\) to an estimate \(\hat y_{hq}\), without any task-specific expert.
We formulate UIR as a conditional flow matching problem in pixel space and present PixRestore, whose network framework is shown in Fig.~\ref{fig:framework}. A VAE-free pixel DiT learns the conditional flow directly on RGB pixels, avoiding the lossy compression of a latent autoencoder (Sec.~\ref{sec:overview}). To capture both degradation and semantic cues, we use a vision encoder to extract multi-layer dense features from the LQ image, and use an adaptive layer router to predict per-layer weights $p_l$. These weights fuse the features into a single representation, which is injected into the DiT blocks by cross-attention. In addition, the predicted weights enable hierarchical visual supervision, detailed in Sec.~\ref{sec:dino}. Finally, for efficient inference, we finetune a single-step generator from the multi-step model via DINO-based adversarial objectives \cite{sun2023improving} (Sec.~\ref{sec:distill}). 
% which consists of three components. (1) A VAE-free pixel DiT learns a conditional flow directly on RGB pixels, avoiding the lossy compression of a latent autoencoder. (2) An adaptive hierarchical visual guidance module extracts multi-layer dense features from \(y_{lq}\) by a vision encoder and predicts layer weights from LQ features by an adaptive layer router, providing both dense conditioning and hierarchical supervision. During training the HQ branch additionally supplies a layer-wise teacher prior; it is removed at inference. (3) A single-step distillation converts the multi-step flow model into a one-step generator via DINO-based adversarial objectives. 

\subsection{Pixel-space Restoration Diffusion Model}
\label{sec:overview}

Instead of encoding images into a compressed latent space~\cite{fapeir}, PixRestore operates directly on RGB pixels.  Given the HQ target \(y_{hq}\), we adopt the linear interpolation path \(x_t=(1-t)\,y_{hq}+t\epsilon\) with \(\epsilon\sim\mathcal N(0,I)\) and \(t\sim\mathcal U(0,1)\). The restoration DiT \(f_\theta\) predicts the clean image as:
\begin{equation}
    \hat{y}_{hq}
    =f_\theta\!\bigl([\,y_{lq};x_t\,],\,t,\,\mathcal F(y_{lq})\bigr),
    \label{eq:predict}
\end{equation}
where \([\,\cdot\,;\,\cdot\,]\) denotes channel-wise concatenation and \(\mathcal F(y_{lq})\) denotes the multi-layer DINO features of the LQ input. %Predicting image rather than the velocity makes pixel-space auxiliary losses be attached directly to \(\hat y_{hq}\). 
Following JiT \cite{jit}, the flow-matching objective is defined on the velocity. Therefore, we recover the velocity from the output by $v_t=(x_t-y_{hq})/t$ and $\hat v_t=(x_t-\hat{y}_{hq})/t$. To prevent division-by-zero as $t \to 0$, we clip the denominator of $1/t$ (by default, at $0.05$) during computation. The flow matching loss is $\mathcal L_{\mathrm{flow}}=\|\hat v_t-v_t\|_2^2$. 
% \begin{equation}
% .
%     \label{eq:flow}
% \end{equation}

\begin{table*}[t]
  \caption{Latent diffusion vs. pixel diffusion under the same UIR training/test setting. The results are averaged over 8 restoration tasks. Pixel-space modeling provides a better overall trade-off in fidelity, perceptual quality, parameter count, and inference speed.}
  \vspace{-3mm}
  \centering
  \scriptsize
  \begin{tabular}{lccccccc}
    \toprule
    Model & {VAE} & {Params(M)} & {Inf Time-DM (ms/step)} & {Inf Time-VAE (ms)} & PSNR (dB)\,$\uparrow$ & { LPIPS}\,$\downarrow$ & { MUSIQ}\,$\uparrow$ \\
    \midrule
    Latent DiT-S  & SD2VAE (f8c4, ps1) & 106.41 & 25 & 78 &  22.10 & 	0.2483  & 50.38  \\
    Latent DiT-S  & FluxVAE (f8c16, ps1) & 106.59  & 25  & 78  & 22.63 & 	0.2109  & 50.86  \\
    Latent DiT-S  & QwenVAE (f8c16, ps1) 
    & 67.37 & 25 & 41 & 22.80 & 	0.2181  & 51.87  \\
    \rowcolor{lotuspink} \textbf{Pixel DiT-S}    & \textbf{None (ps8)}   & \textbf{23.41} & 25 & 0 & \textbf{26.62} & 	\textbf{0.1593} & 	\textbf{54.32}  \\
    \bottomrule
  \end{tabular}
  \label{tab:pixel_vs_latent}
  %\vspace{-4mm}
\end{table*}

The concatenated input \([y_{lq};x_t]\) has six channels. A patch embedding partitions the full-resolution pixel grid into tokens, shown in Fig. \ref{fig:framework}. In this way, a relatively large patch size keeps the token sequence tractable without a VAE. Each DiT block consists of RMSNorm, QK-normalized attention, and rotary positional embeddings \cite{vavae}. A single shared timestep block produces AdaLN modulation parameters for all Transformer blocks~\cite{Pixart-alpha}. Multi-layer DINO features are injected via cross-attention in each block. During inference, the model integrates the predicted velocity from Gaussian noise using an Euler solver.

The overall training objective combines the flow loss with two auxiliary terms from the DINO module:
\begin{equation}
    \mathcal L
    =\mathcal L_{\mathrm{flow}}
    +\lambda_{\mathrm{wpred}}\mathcal L_{\mathrm{wpred}}
    +\lambda_{\mathrm{feat}}\mathcal L_{\mathrm{feat}},
    \label{eq:total}
\end{equation}
where \(\mathcal L_{\mathrm{wpred}}\) supervises the adaptive layer router and \(\mathcal L_{\mathrm{feat}}\) enforces hierarchical feature fidelity. The two losses will be discussed and defined in Sec.~\ref{sec:dino}.

\noindent\textbf{Pixel Space vs. Latent Space.}
Latent DiTs use a VAE to reduce computational complexity before diffusion, but this compression can sacrifice small structures and fine textures. Pixel DiT instead operates on RGB pixels and keeps their spatial structure through patchification, better matching IR tasks, where the output must stay faithful to the input. To verify this, we compare the same DiT-S in the pixel space against latent space built on three widely used VAEs, \textit{i.e.}, SD2VAE~\cite{sd2.1}, FluxVAE~\cite{FLUX} and QwenVAE~\cite{Qwen-Image-Edit}, under identical training and test settings, including training data, resolution, diffusion architecture, optimizer, and sampler. The pixel DiT model uses a patch size of 8 to match the latent resolution. 

As shown in Table~\ref{tab:pixel_vs_latent}, pixel DiT beats all latent DiTs across all metrics (26.62 dB PSNR, 0.1593 LPIPS, 54.32 MUSIQ), significantly outperforming the best baseline with QwenVAE. Removing the VAE also reduces cost and latency (41--78\,ms). Using only 23.41M parameters without VAE, our pixel DiT design reproduces more faithful image details, \textit{indicating that pixel-space modeling can better match the requirements of UIR}. Detailed training and test settings, as well as per-degradation analysis, are provided in the \textbf{Appendix}.

\subsection{Adaptive Hierarchical Visual Guidance}
\label{sec:dino}

\noindent\textbf{Visual Foundation Prior}.
To handle different types of degradations in the input LQ image, we introduce a frozen vision foundation encoder to provide dense visual cues for UIR. Specifically, we adopt DINOv2 \cite{dinov2} for this purpose because its self-distillation pretraining can produce dense, spatially precise tokens that preserve fine structure and texture while being semantically discriminative. We validate this choice with an experimental comparison against other encoders (CLIP~\cite{CLIP}, MAE~\cite{MAE}, SigLIP~\cite{siglip}, DINOv2~\cite{dinov2}) under the same pixel DiT setting, where DINOv2 performs the best on almost all metrics. The experiment details are in the \textbf{Appendix}.

\begin{figure*}[t]
  \centering
  \includegraphics[width=\linewidth]{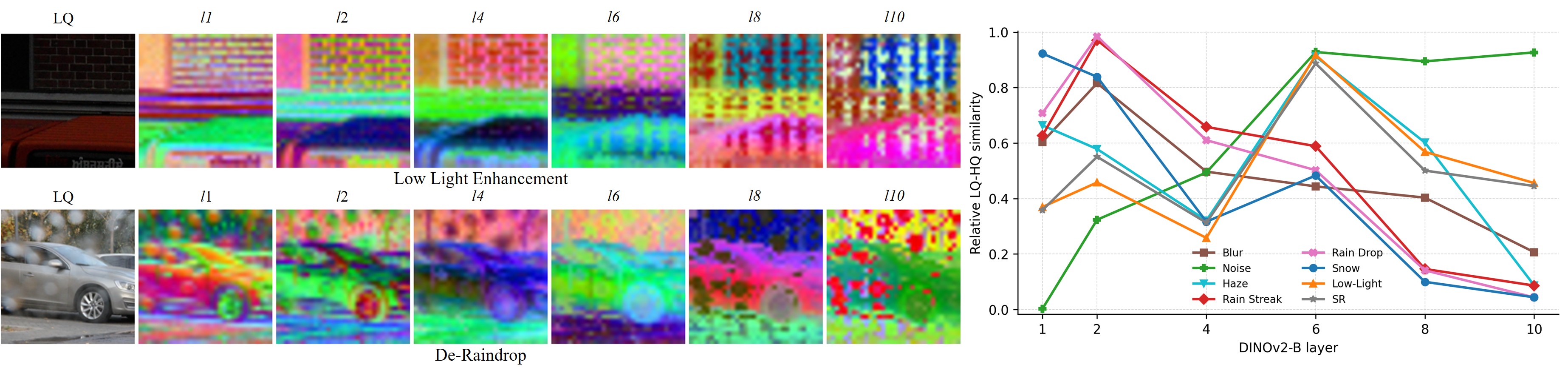}
  \vspace{-8mm}
  \caption{Motivation of adaptive hierarchical visual guidance. \textbf{Left:} Per-layer DINO feature visualizations for low-light enhancement and de-raindrop. Shallow layers preserve local structures and details, while deeper layers encode global semantics. \textbf{Right:} LQ--HQ feature similarity across DINOv2-B layers for eight types of degradations. We see that different layers are sensitive to different degradations.}
  \label{fig:motivation}
  \vspace{-4mm}
\end{figure*}

\noindent\textbf{Similarity-Guided Adaptive Layer Router}.
As illustrated in Fig.~\ref{fig:motivation} (left), different DINO layers carry complementary visual cues: shallow layers ($l_1$--$l_2$) preserve local structures such as edges and textures for detailed reconstruction, while deeper layers ($l_8$--$l_{10}$) encode global semantics for degradation and content discrimination. Layer sensitivity is also degradation-dependent. As shown in Fig.~\ref{fig:motivation} (right), which presents the LQ-HQ feature similarity, no single layer is sensitive to all degradations. In particular, shallow layers are sensitive to rain, blur, and snow degradations; deeper layers are sensitive to noise; and middle layers are sensitive to low-light, SR, and haze. Using a fixed layer or a uniform average over layers is thus suboptimal. We therefore train a lightweight module to predict per-image layer weights, supervised by paired LQ-HQ similarity.

For each layer \(l\), we measure how much the LQ features retain the HQ content by averaging a cosine similarity and a normalized $L_2$-distance similarity over the projected LQ and HQ patch tokens: $s_l=\tfrac{1}{2}\big(s_l^{\mathrm{cos}}+s_l^{\mathrm{dist}}\big)$. A larger \(s_l\) means that layer \(l\) is more reliable under the observed degradation, so it should have a larger weight:
\begin{equation}
    q_l=\frac{\exp(s_l)}{\sum_{k\in\mathcal L}\exp(s_k)}.
    \label{eq:qweight}
\end{equation}

Computing \(q_l\) requires the HQ image, which is unavailable at inference time. Thus, we train a lightweight predictor \(\rho_\psi\) to estimate the weights from the LQ image and features: $p_l=\operatorname{softmax}\!\big(\rho_\psi([y_{lq},{U_l}])\big), {l\in\mathcal L}$, where \(U_l\) denotes the projected LQ features of different DINO layers. The cross-entropy loss is used to supervise the training:
\begin{equation}
    \mathcal L_{\mathrm{wpred}}=-\sum_{l\in\mathcal L}q_l\log p_l.
    \label{eq:prior}
\end{equation}
The predictor thus learns which DINO layers are more reliable for each input, instead of relying on a uniform mixture.

\noindent\textbf{Adaptive Conditioning and Hierarchical Supervision}.
The projected LQ features are fused using the predicted weights as follows:
\begin{equation}
    U_{\mathrm{fuse}}=\sum_{l\in\mathcal L}p_l U_l,
    \label{eq:fuse}
\end{equation}
which are then injected into each DiT block via cross-attention, with image tokens as queries and \(U_{\mathrm{fuse}}\) as keys and values.
To emphasize layers where the LQ features differ most from the HQ features, we further introduce a hierarchical feature supervision loss:
\begin{equation}
        \qquad
    \mathcal L_{\mathrm{feat}}
    =\sum_{l\in\mathcal L}r_l\ell_l^{\mathrm{feat}}, \
    r_l
    =\frac{\exp((1-s_l))}
    {\sum_{k\in\mathcal L}\exp((1-s_k))},
    \label{eq:featloss}
\end{equation}
where \(\ell_l^{\mathrm{feat}}\) is the cosine similarity loss between the HQ features and the restored-output features at layer \(l\). The weights \(q_l\) (see Eq. ~\eqref{eq:qweight}) and \(r_l\) are complementary, \textit{i.e.}, \(q_l\) selects reliable content features for conditioning, while \(r_l\) focuses supervision on layers that need stronger restoration. With \(\mathcal L_{\mathrm{wpred}}\) and \(\mathcal L_{\mathrm{feat}}\) defined above, the complete multi-step objective is given by Eq.~\eqref{eq:total}, where both auxiliary weights are set to \(0.5\), and all layer features are channel-wise RMS normalized before projection.

\subsection{Single-step Finetuning}
\label{sec:distill}

The multi-step PixRestore model requires iterative sampling. We therefore finetune it into a single-step generator for efficient UIR. We initialize the student from the pretrained multi-step teacher and fix the flow time to \(t=1\) so that the generator can predict the clean image in one forward pass from pure Gaussian noise $\epsilon$, \textit{i.e.}, $\hat{y}_{hq}=f_\theta\!\left([\,y_{lq};\epsilon\,],\,t=1,\,\mathcal F(y_{lq})\right)$.

Since one-step generation may lose fine textures, we add a DINO-based adversarial objective. Reusing the same frozen encoder and layers, a lightweight multi-layer discriminator \(D\) distinguishes the restored features \(\mathcal F(\hat y_{hq})\) from the HQ features \(\mathcal F(y_{hq})\) with an independent head per layer. The discriminator is trained with the standard binary cross-entropy loss to classify \(F_l^{y_{hq}}\) as real and \(F_l^{\hat y_{hq}}\) as fake:
\begin{equation}
    \mathcal L_{D}
    =\frac{1}{|\mathcal L|}\sum_{l\in\mathcal L}
    \Big[
    \ell_{\mathrm{bce}}\!\left(D_l\!\left(F_l^{y_{hq}}\right),1\right)
    +
    \ell_{\mathrm{bce}}\!\left(D_l\!\left(F_l^{\hat y_{hq}}\right),0\right)
    \Big].
\end{equation}
The generator is optimized to fool \(D\):
\begin{equation}
    \mathcal L_{\mathrm{adv}}
    =\frac{1}{|\mathcal L|}\sum_{l\in\mathcal L}
    \ell_{\mathrm{bce}}\!\left(D_l\!\left(F_l^{\hat y_{hq}}\right),1\right).
    \label{eq:adv}
\end{equation}
The generator and discriminator are updated alternately. With \(t=1\), the single-step objective is:
\begin{equation}
    \mathcal L
    =\mathcal L_{\mathrm{flow}}
    +\lambda_{\mathrm{wpred}}\mathcal L_{\mathrm{wpred}}
    +\lambda_{\mathrm{feat}}\mathcal L_{\mathrm{feat}}
    +\lambda_{\mathrm{adv}}\mathcal L_{\mathrm{adv}},
    \label{eq:singlestep}
\end{equation}
where all auxiliary weights are set to \(0.5\). Operating on frozen DINO tokens rather than raw pixels, the discriminator shares the conditioning prior and adds little overhead. During inference, PixRestore restores an image from one noise sample with the LQ condition in a single step.

%==========================================================================
% \lipsum[7]

%==========================================================================
\section{Experiments}
\subsection{Experimental Setup}
\label{exp-set}
% TODO: add experimental setup, benchmarks, comparisons, and ablations.

\noindent\textbf{Training and Test Datasets.} We build a training corpus of about 2.83M images covering eight restoration tasks (deblur, dehaze, denoise, de-rainstreak, de-raindrop, desnow, low-light enhancement, and super-resolution (SR)), with samples drawn with equal probability during training. 

We evaluate PixRestore under two complementary settings. The first uses \textbf{public benchmarks with paired GT} for fidelity and perceptual evaluation: GoPro \cite{GoPro} and UHD-blur \cite{uhdblurhaze} (deblur), RESIDE-6K \cite{RESIDE} and UHD-Haze \cite{uhdblurhaze} (dehaze), DIV2K \cite{div2k} (Gaussian noise) and PolyU \cite{polyunoise} (denoise), RainDS-real \cite{rainds} and RealRain-1k \cite{realrain} (de-rainstreak), RainDS-real \cite{rainds} and UAV-Rain1k \cite{UAV-Rain1k} (de-raindrop), UHD-LL \cite{uhdll} and LOLdataset \cite{LoL} (low-light), WeatherBench \cite{guan2025weatherbench} (desnow), and RealSR \cite{realsr} and ScreenSR \cite{vosr} (SR), all center-cropped to 512 for testing. The second is a \textbf{real-world test set without GT}, with 100 LQ images per degradation (deblur, dehaze, de-rainstreak, de-raindrop, desnow, low-light) from diverse sources \cite{jarvisir2025,real-Desnownet,real-nturain_real}. Detailed training and testing data are given in the \textbf{Appendix}.
% {Due to the page limit, detailed public benchmaare given in \textbf{Appendix}, where PixRestore shows strong generalization to complex real-world degradations.}

\noindent\textbf{Compared Methods.}
We compare with representative UIR methods, including the regression-based PromptIR \cite{PromptIR} and diffusion-based DA-CLIP \cite{DA-CLIP}, DiffUIR \cite{diffuir}, FoundIR \cite{li2025foundir}, FoundIR-v2 \cite{foundirv2}, UniRestore \cite{unirestore}, Flux-IR \cite{fluxir} and FAPE-IR \cite{fapeir}. {For fair comparison and to isolate the effect of training data, we evaluate both the official checkpoints and retrained versions of the major baselines on our dataset.}

\noindent\textbf{PixRestore Model Settings.}
PixRestore adopts LightningDiT \cite{vavae} as the DiT backbone and DINOv2 \cite{dinov2} as the vision encoder. Each Transformer block follows the LightningDiT design and contains a multi-head self-attention module, a cross-attention module, and a feed-forward network. A single shared timestep block produces AdaLN modulation parameters for all Transformer blocks~\cite{Pixart-alpha}. We provide four variants of the model with different backbone sizes. All variants use a patch size of 8 in pixel patchification. During training, the DINOv2 model is frozen and only the diffusion model is trained.  Unless otherwise specified, ``PixRestore'' refers to the PixRestore-S variant in the paper.

\begin{itemize}
    \item \textbf{PixRestore-S} uses a LightningDiT-S backbone with a hidden dimension of 384. It contains 12 Transformer blocks, each with 6 attention heads. DINOv2-S is used as the vision encoder.
    \item \textbf{PixRestore-B} uses a LightningDiT-B backbone with a hidden dimension of 768. It contains 12 Transformer blocks, each with 12 attention heads. DINOv2-B is used as the vision encoder.
    \item \textbf{PixRestore-L} uses a LightningDiT-L backbone with a hidden dimension of 1024. It contains 24 Transformer blocks, each with 16 attention heads. DINOv2-L is used as the vision encoder.
    \item \textbf{PixRestore-XL} uses a LightningDiT-XL backbone with a hidden dimension of 1152. It contains 28 Transformer blocks, each with 16 attention heads. Since DINOv2 does not provide an XL version, we use DINOv2-L as the vision encoder.
\end{itemize}

\noindent\textbf{Training Details.}
We train the multi-step model for 250K iterations, and then finetune it into a one-step model for an additional 100K iterations, using AdamW with a learning rate of \(1\times10^{-4}\) and a batch size of 16 on \(512\times512\) image crops across 8 NVIDIA A800 GPUs. DINO features are extracted from six layers evenly distributed throughout the encoder. 
% The model variants denoted by S, B, L, and XL use DINO-S, DINO-B, DINO-L, and DINO-L, respectively. Unless otherwise specified, we adopt the default PixRestore setting with DINO-S and a patch size of 8. 

\begin{figure}[t]
  \centering
  \includegraphics[width=0.65\linewidth]{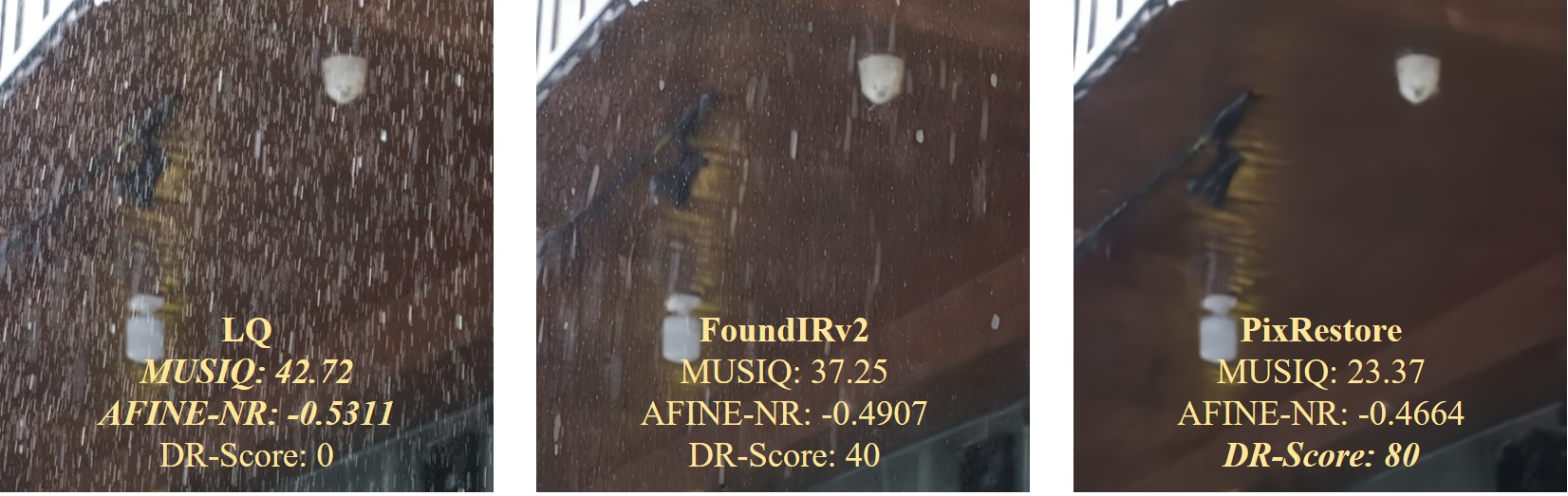}
  % \vspace{-4mm}
  \caption{No-reference quality metrics do not reliably reflect degradation removal. Here PixRestore removes the rainstreaks best, yet MUSIQ and AFINE-NR rank it worst, favoring the LQ input and the degradation-preserving output of FoundIR-v2, while our VLM-based \textbf{DR-Score} demonstrates strong alignment with human perceptual judgments.}
  \label{fig:vlmscore}
  % \vspace{-4mm}
\end{figure}

\begin{table*}[t]
  \caption{Quantitative comparison on public benchmarks. The best and second-best results are highlighted in {\textcolor{red}{\textbf{red}}} and {\textcolor{blue}{\textit{blue italic}}}, respectively. Methods marked with $^*$ are retrained using the same training dataset as ours. Metrics: PSNR$\uparrow$, SSIM$\uparrow$, LPIPS$\downarrow$, DISTS$\downarrow$, DR-Score (degradation-removal score)$\uparrow$.}
  \label{tab:synthetic_uir_comparison}
  \vspace{-2mm}
  \centering
  \scriptsize
  \setlength{\tabcolsep}{1.2pt}
  \renewcommand{\arraystretch}{0.90}
  \setlength{\aboverulesep}{0.3pt}
  \setlength{\belowrulesep}{0.3pt}
  \resizebox{\textwidth}{!}{%
  \begin{tabular}{l*{20}{c}}
    \toprule
    Method & \multicolumn{5}{c}{De-rainstreak} & \multicolumn{5}{c}{Denoise} & \multicolumn{5}{c}{Deblur} & \multicolumn{5}{c}{De-raindrop} \\
    \cmidrule(lr){2-6} \cmidrule(lr){7-11} \cmidrule(lr){12-16} \cmidrule(lr){17-21}
     & PSNR & SSIM & LPIPS & DISTS & DR-Score & PSNR & SSIM & LPIPS & DISTS & DR-Score & PSNR & SSIM & LPIPS & DISTS & DR-Score & PSNR & SSIM & LPIPS & DISTS & DR-Score \\
    \midrule
     PromptIR & 24.50 & 0.7507 & 0.3529 & 0.2381 & 31.25 & 32.54 & 0.9029 & 0.2459 & 0.1604 & 55.35 & 23.82 & 0.7584 & 0.3131 & 0.2195 & 27.23 & 18.81 & 0.6962 & 0.3390 & 0.1885 & 24.86 \\
    PromptIR$^*$ & 28.43 & 0.8389 & 0.2659 & 0.1904 & 49.99 & \textcolor{blue}{\textit{35.45}} & \textcolor{blue}{\textit{0.9377}} & 0.1104 & 0.1065 & 79.33 & 29.10 & \textcolor{blue}{\textit{0.8474}} & 0.2051 & 0.1560 & 55.56 & 23.68 & \textcolor{blue}{\textit{0.8005}} & 0.2236 & 0.1326 & 51.74 \\
    DiffUIR & 24.51 & 0.7618 & 0.3484 & 0.2248 & 40.13 & 26.61 & 0.8380 & 0.3103 & 0.2056 & 59.52 & 27.86 & 0.8200 & 0.2258 & 0.1699 & 44.09 & 18.85 & 0.7004 & 0.3455 & 0.1905 & 26.50 \\
    UniRestore & 22.50 & 0.7291 & 0.4223 & 0.2689 & 34.31 & 31.75 & 0.8969 & 0.2336 & 0.1669 & 61.20 & 23.87 & 0.7205 & 0.2405 & 0.1716 & 46.94 & 18.50 & 0.6377 & 0.4114 & 0.2285 & 27.90 \\
    DA-CLIP & 24.50 & 0.7560 & 0.3347 & 0.2176 & 43.09 & 27.35 & 0.8245 & 0.2459 & 0.1764 & 64.12 & 27.40 & 0.8113 & 0.1641 & 0.1303 & 55.00 & 20.12 & 0.6988 & 0.2715 & 0.1512 & 55.20 \\
    DA-CLIP$^*$ & 31.61 & 0.8606 & 0.1048 & 0.0885 & 78.72 & 33.97 & 0.8728 & 0.1462 & 0.1166 & 75.14 & 28.29 & 0.8228 & 0.1474 & 0.1148 & 58.38 & 23.58 & 0.7916 & 0.1325 & \textcolor{blue}{\textit{0.0852}} & 79.64 \\
    FoundIR & 26.87 & 0.8263 & 0.2454 & 0.1799 & 52.37 & 32.46 & 0.7925 & 0.2794 & 0.1641 & 55.52 & 27.29 & 0.8046 & 0.2430 & 0.1781 & 39.12 & 18.87 & 0.6972 & 0.3586 & 0.2017 & 26.21 \\
    FoundIR$^*$ & \textcolor{blue}{\textit{32.36}} & \textcolor{blue}{\textit{0.8888}} & 0.1593 & 0.1203 & 71.12 & \textcolor{red}{\textbf{36.16}} & \textcolor{red}{\textbf{0.9415}} & 0.1051 & 0.1245 & 80.86 & \textcolor{blue}{\textit{29.14}} & 0.8466 & 0.1986 & 0.1503 & 49.86 & 24.40 & \textcolor{red}{\textbf{0.8218}} & 0.1942 & 0.1146 & 65.00 \\
    Flux-IR & 20.98 & 0.6233 & 0.4625 & 0.2861 & 26.04 & 25.71 & 0.6842 & 0.4197 & 0.2367 & 54.86 & 23.51 & 0.6745 & 0.2746 & 0.1851 & 58.70 & 18.94 & 0.6459 & 0.3224 & 0.1774 & 39.72 \\
    Flux-IR$^*$ & 21.01 & 0.6344 & 0.4511 & 0.2822 & 31.04 & 26.80 & 0.7410 & 0.3789 & 0.2051 & 37.05 & 22.15 & 0.6446 & 0.2818 & 0.2036 & \textcolor{blue}{\textit{73.18}} & 17.80 & 0.5785 & 0.3266 & 0.2075 & 71.26 \\
     FoundIR-v2 & 23.17 & 0.6652 & 0.3659 & 0.2218 & 57.26 & 26.87 & 0.7481 & 0.2785 & 0.1946 & 74.47 & 24.40 & 0.7079 & 0.2203 & 0.1566 & 73.23 & 19.63 & 0.5459 & 0.2792 & 0.1563 & 55.48 \\
    FoundIR-v2$^*$ & 27.85 & 0.7598 & 0.1773 & 0.1322 & 81.03 & 28.17 & 0.7497 & 0.2889 & 0.1890 & 75.01 & 24.98 & 0.7351 & 0.1914 & 0.1323 & \textcolor{red}{\textbf{77.12}} & 20.81 & 0.5461 & 0.2300 & 0.1294 & 80.28 \\
    FAPE-IR & 27.51 & 0.8226 & 0.2319 & 0.1679 & 71.09 & 32.99 & 0.9088 & 0.1238 & 0.1113 & 79.31 & 26.80 & 0.7895 & 0.2098 & 0.1556 & 50.72 & 21.39 & 0.6822 & 0.2219 & 0.1316 & 71.77 \\
    FAPE-IR$^*$ & 31.91 & 0.8695 & 0.0903 & \textcolor{red}{\textbf{0.0762}} & \textcolor{blue}{\textit{84.64}} & 34.22 & 0.9172 & 0.0741 & 0.0750 & \textcolor{blue}{\textit{82.56}} & 27.96 & 0.8207 & 0.1577 & 0.1094 & 66.75 & 23.88 & 0.7223 & 0.1555 & 0.0967 & 81.91 \\
    \rowcolor{lotuspink} \textbf{PixRestore} & 32.28 & 0.8847 & \textcolor{blue}{\textit{0.0902}} & 0.0905 & 82.75 & 34.87 & 0.9336 & \textcolor{blue}{\textit{0.0624}} & \textcolor{blue}{\textit{0.0735}} & 81.89 & 28.32 & 0.8284 & \textcolor{blue}{\textit{0.1201}} & \textcolor{blue}{\textit{0.0940}} & 70.19 & \textcolor{blue}{\textit{24.48}} & 0.7755 & \textcolor{blue}{\textit{0.1258}} & 0.0882 & \textcolor{blue}{\textit{82.20}} \\
    \rowcolor{lotuspink} \textbf{PixRestore-B} & \textcolor{red}{\textbf{32.85}} & \textcolor{red}{\textbf{0.8898}} & \textcolor{red}{\textbf{0.0767}} & \textcolor{blue}{\textit{0.0817}} & \textcolor{red}{\textbf{84.72}} & 34.62 & 0.9356 & \textcolor{red}{\textbf{0.0564}} & \textcolor{red}{\textbf{0.0706}} & \textcolor{red}{\textbf{83.70}} & \textcolor{red}{\textbf{29.23}} & \textcolor{red}{\textbf{0.8511}} & \textcolor{red}{\textbf{0.1051}} & \textcolor{red}{\textbf{0.0835}} & 72.69 & \textcolor{red}{\textbf{25.21}} & 0.7976 & \textcolor{red}{\textbf{0.1087}} & \textcolor{red}{\textbf{0.0763}} & \textcolor{red}{\textbf{84.77}} \\
     % \rowcolor{lotuspink} \textbf{PixRestore-L} & \textcolor{red}{\textbf{33.03}} & \textcolor{red}{\textbf{0.8912}} & \textcolor{red}{\textbf{0.0697}} & \textcolor{blue}{\textit{0.0799}} & \textcolor{red}{\textbf{84.71}} & 34.63 & 0.9349 & \textcolor{red}{\textbf{0.0557}} & \textcolor{blue}{\textit{0.0738}} & \textcolor{red}{\textbf{82.75}} & \textcolor{red}{\textbf{29.87}} & \textcolor{red}{\textbf{0.8647}} & \textcolor{red}{\textbf{0.0952}} & \textcolor{red}{\textbf{0.0774}} & \textcolor{blue}{\textit{72.50}} & \textcolor{red}{\textbf{25.52}} & \textcolor{blue}{\textit{0.8067}} & \textcolor{red}{\textbf{0.1003}} & \textcolor{red}{\textbf{0.0703}} & \textcolor{red}{\textbf{83.52}} \\
    \midrule[\heavyrulewidth]
    \midrule[\heavyrulewidth]
    Method & \multicolumn{5}{c}{Desnow} & \multicolumn{5}{c}{Dehaze} & \multicolumn{5}{c}{Low-light enhancement} & \multicolumn{5}{c}{Super-resolution} \\
    \cmidrule(lr){2-6} \cmidrule(lr){7-11} \cmidrule(lr){12-16} \cmidrule(lr){17-21}
     & PSNR & SSIM & LPIPS & DISTS & DR-Score & PSNR & SSIM & LPIPS & DISTS & DR-Score & PSNR & SSIM & LPIPS & DISTS & DR-Score & PSNR & SSIM & LPIPS & DISTS & DR-Score \\
    \midrule
     PromptIR & 22.26 & 0.7939 & 0.2452 & 0.1672 & 24.35 & 21.34 & 0.8803 & 0.1426 & 0.1012 & 61.68 & 10.49 & 0.4781 & 0.5410 & 0.3765 & 27.87 & 24.26 & 0.7372 & 0.4394 & 0.2510 & 33.48 \\
    PromptIR$^*$ & 29.32 & 0.8532 & 0.1818 & 0.1402 & 72.39 & 21.25 & 0.8931 & 0.1348 & 0.0885 & 67.89 & 17.91 & 0.6881 & 0.3564 & 0.2512 & 58.58 & \textcolor{blue}{\textit{27.59}} & \textcolor{red}{\textbf{0.7968}} & 0.2839 & 0.2208 & 55.70 \\
    DiffUIR & 22.95 & 0.7948 & 0.2392 & 0.1667 & 25.00 & 20.41 & 0.8656 & 0.1640 & 0.1175 & 57.19 & 21.72 & 0.7082 & 0.3819 & 0.2292 & 62.93 & 26.54 & 0.7582 & 0.3866 & 0.2378 & 45.81  \\
    UniRestore & 22.33 & 0.7863 & 0.2464 & 0.1771 & 26.12 & 20.11 & 0.8469 & 0.2122 & 0.1359 & 67.33 & 10.94 & 0.5188 & 0.5006 & 0.3165 & 38.38 & 24.80 & 0.7543 & 0.3548 & 0.2284 & 60.49 \\
    DA-CLIP & 23.60 & 0.7971 & 0.2221 & 0.1558 & 35.53 & 22.96 & 0.8751 & 0.1311 & 0.0885 & 70.33 & 22.25 & 0.7915 & 0.2469 & 0.1622 & 73.29 & 23.73 & 0.6789 & 0.3772 & 0.2367 & 46.43 \\
    DA-CLIP$^*$ & 28.31 & 0.8282 & 0.1360 & 0.1021 & 81.48 & 21.92 & 0.8779 & 0.1459 & 0.1077 & 58.58 & 18.01 & 0.8185 & 0.2129 & 0.1614 & 72.79 & 26.49 & 0.7585 & 0.2341 & 0.1763 & 65.25 \\
    FoundIR & 23.03 & 0.7999 & 0.2406 & 0.1630 & 24.68 & 15.07 & 0.7901 & 0.2582 & 0.1919 & 33.17 & 15.34 & 0.7473 & 0.3034 & 0.2132 & 62.24 & 25.85 & 0.7399 & 0.4280 & 0.2492 & 29.70 \\
    FoundIR$^*$ & 29.82 & 0.8678 & 0.1524 & 0.1224 & 73.00 & 20.99 & 0.8903 & 0.1335 & 0.0962 & 64.81 & 23.34 & \textcolor{red}{\textbf{0.9026}} & 0.1792 & 0.1439 & 80.41 & \textcolor{red}{\textbf{27.65}} & \textcolor{blue}{\textit{0.7952}} & 0.2883 & 0.2251 & 64.69 \\
    Flux-IR & 21.74 & 0.7231 & 0.3434 & 0.2204 & 26.80 & 14.73 & 0.7599 & 0.2976 & 0.1986 & 46.13 & 18.85 & 0.7022 & 0.3557 & 0.1998 & 62.16 & 22.49 & 0.6541 & 0.2903 & 0.2064 & 77.19 \\
    Flux-IR$^*$ & 21.79 & 0.6831 & 0.3023 & 0.1959 & 51.21 & 15.96 & 0.8061 & 0.2315 & 0.1578 & 52.88 & 18.40 & 0.7198 & 0.3386 & 0.2010 & 61.45 & 20.87 & 0.5853 & 0.3569 & 0.2519 & 74.89 \\
     FoundIR-v2 & 24.72 & 0.7347 & 0.2513 & 0.1671 & 67.47 & 19.06 & 0.7705 & 0.1928 & 0.1337 & 62.79 & 17.17 & 0.7448 & 0.3132 & 0.2027 & 72.49 & 23.85 & 0.6661 & 0.2959 & 0.1974 & 76.25 \\
    FoundIR-v2$^*$ & 26.28 & 0.7690 & 0.1824 & 0.1311 & 82.52 & 19.10 & 0.7645 & 0.1933 & 0.1282 & 63.22 & 17.04 & 0.7307 & 0.3188 & 0.2101 & 68.16 & 23.79 & 0.6530 & 0.2493 & 0.1682 & \textcolor{red}{\textbf{78.87}} \\
    FAPE-IR & 26.02 & 0.8191 & 0.1759 & 0.1189 & 60.36 & 25.28 & 0.9007 & 0.0960 & 0.0650 & 75.76 & 19.46 & 0.7629 & 0.2580 & 0.1838 & 58.29 & 26.55 & 0.7732 & 0.2816 & 0.1977 & 68.92 \\
    FAPE-IR$^*$ & 30.19 & 0.8676 & 0.1136 & 0.0849 & 83.66 & 25.34 & 0.9056 & 0.0927 & \textcolor{blue}{\textit{0.0645}} & \textcolor{blue}{\textit{76.59}} & \textcolor{red}{\textbf{26.04}} & \textcolor{blue}{\textit{0.9014}} & 0.1370 & 0.1068 & \textcolor{blue}{\textit{82.00}} & 27.48 & 0.7843 & 0.1952 & 0.1433 & 77.81 \\
    \rowcolor{lotuspink} \textbf{PixRestore} & \textcolor{blue}{\textit{31.26}} & \textcolor{blue}{\textit{0.8859}} & \textcolor{blue}{\textit{0.0853}} & \textcolor{blue}{\textit{0.0669}} & \textcolor{blue}{\textit{84.22}} & \textcolor{blue}{\textit{25.46}} & \textcolor{blue}{\textit{0.9142}} & \textcolor{blue}{\textit{0.0896}} & 0.0650 & 74.86 & 25.62 & 0.8894 & \textcolor{blue}{\textit{0.1360}} & \textcolor{blue}{\textit{0.1008}} & 81.58 & 27.01 & 0.7730 & \textcolor{blue}{\textit{0.1736}} & \textcolor{blue}{\textit{0.1358}} & 77.53 \\
    \rowcolor{lotuspink} \textbf{PixRestore-B} & \textcolor{red}{\textbf{31.93}} & \textcolor{red}{\textbf{0.8959}} & \textcolor{red}{\textbf{0.0688}} & \textcolor{red}{\textbf{0.0584}} & \textcolor{red}{\textbf{84.51}} & \textcolor{red}{\textbf{26.64}} & \textcolor{red}{\textbf{0.9228}} & \textcolor{red}{\textbf{0.0789}} & \textcolor{red}{\textbf{0.0574}} & \textcolor{red}{\textbf{76.60}} & \textcolor{blue}{\textit{25.72}} & 0.8935 & \textcolor{red}{\textbf{0.1298}} & \textcolor{red}{\textbf{0.0955}} & \textcolor{red}{\textbf{83.08}} & 27.07 & 0.7773 & \textcolor{red}{\textbf{0.1601}} & \textcolor{red}{\textbf{0.1289}} & \textcolor{blue}{\textit{77.88}} \\
     % \rowcolor{lotuspink} \textbf{PixRestore-L} & \textcolor{red}{\textbf{32.35}} & \textcolor{red}{\textbf{0.9039}} & \textcolor{red}{\textbf{0.0656}} & \textcolor{red}{\textbf{0.0569}} & \textcolor{red}{\textbf{85.28}} & \textcolor{red}{\textbf{26.93}} & \textcolor{red}{\textbf{0.9248}} & \textcolor{red}{\textbf{0.0796}} & \textcolor{red}{\textbf{0.0568}} & \textcolor{blue}{\textit{75.68}} & \textcolor{red}{\textbf{26.10}} & 0.8944 & \textcolor{red}{\textbf{0.1300}} & \textcolor{red}{\textbf{0.0955}} & \textcolor{red}{\textbf{82.49}} & 27.00 & 0.7721 & \textcolor{red}{\textbf{0.1537}} & \textcolor{red}{\textbf{0.1272}} & 77.22 \\
    \bottomrule
  \end{tabular}%
  }
  
\end{table*}

\noindent\textbf{Evaluation.}
We assess fidelity with PSNR and SSIM \cite{ssim} and perceptual quality with LPIPS \cite{lpips} and DISTS \cite{dists}. Degradation removal is a key indicator for UIR, reflecting whether a method truly eliminates the target degradation, especially in real-world cases where only no-reference (NR) metrics apply. However, existing NR metrics (\textit{e.g.}, MUSIQ \cite{musiq} and AFINE-NR \cite{afine}) do not measure it well. As shown in Fig.~\ref{fig:vlmscore}, PixRestore removes rainstreaks most effectively but scores the worst on MUSIQ and AFINE-NR, which favor the LQ input and the degradation-preserving output of FoundIR-v2. We therefore propose \textbf{DR-Score} as an auxiliary diagnostic metric. It uses a vision-language model (VLM, \textit{e.g.}, Gemini-3.1 Pro \cite{gemini3-1pro}) to judge whether the target degradation has been removed. Because VLM outputs can be stochastic, we test each image five times and report the average score. In the \textbf{Appendix}, we detail the prompt design of DR-Score, and demonstrate its strong alignment with human perceptual judgments.
% (PSNR, SSIM \cite{}, LPIPS \cite{}, and DISTS \cite{}). also report perceptual quality for reference with no-reference metrics (MUSIQ \cite{} and AFINE-NR \cite{}).
% We introduce DR-Score as an auxiliary diagnostic metric for degradation removal and validate it against human judgments.

\begin{figure}[t]
  \centering
  \includegraphics[width=0.9\textwidth]{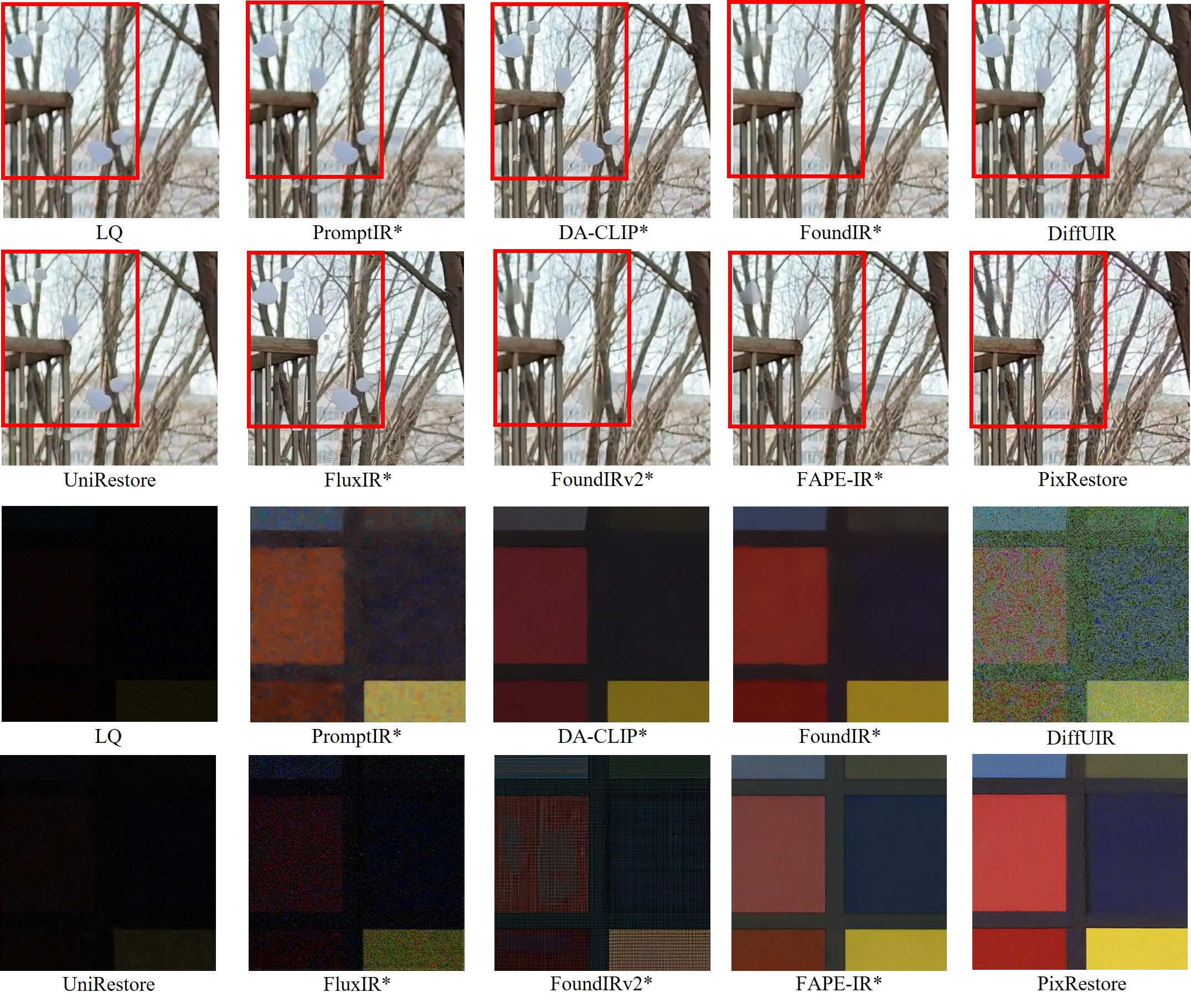}
  % \vspace{-7mm}
  \caption{Visual comparisons on desnow (top) and low-light enhancement (bottom). PixRestore removes degradations effectively and recovers more faithful details and colors.}
  \label{fig:syncom}
  % \vspace{-4mm}
\end{figure}

\begin{table}[htbp]
  \caption{Complexity comparison of different methods. ``NFE'' denotes number of evaluations. All values are measured with input $1\times3\times512\times512$ on a single NVIDIA A800 GPU, with 5 warmup iterations and averaged over 100 runs.}
  \vspace{-3mm}
  \label{tab:complexity}
  \centering
  \scriptsize
  \setlength{\tabcolsep}{6pt}
  \renewcommand{\arraystretch}{1.15}
  \setlength{\aboverulesep}{0.3pt}
  \setlength{\belowrulesep}{0.3pt}
  \begin{tabular}{lcccc}
    \toprule
    Method & NFE & Params (M) & FLOPs (G) & Latency (ms) \\
    \midrule
    PromptIR & - &  35.59 & 1382 & 334 \\
    DiffUIR   & 4 & 36.26 & 3839 & 777 \\
    DA-CLIP   & 100 & 231.76 & 112927 & 18071 \\
    FoundIR   & 4 & 36.26 & 3839 & 777 \\
    UniRestore & 1& 1071.20 & 5255 & 158 \\
    FoundIR-v2 & 20 & 16910.09 & 119334 & 18293 \\
    Flux-IR   &  21 & 17698.23 & 795290 & 5790 \\
    FAPE-IR  & 1 & 21575.43 & 64758 & 1011 \\
    \rowcolor{lotuspink} \textbf{PixRestore} & 1& 53.70 & 658 & 44 \\
    \rowcolor{lotuspink} \textbf{PixRestore-B} & 1 & 210.89 & 1842 & 79 \\
    % \rowcolor{lotuspink} \textbf{PixRestore-L} & 726.10 & 5761 & 201 \\
    \bottomrule
  \end{tabular}
\end{table}

\subsection{Public Benchmark Results}

% \noindent\textbf{Quantitative and qualitative comparisons.} 
Table \ref{tab:synthetic_uir_comparison} compares the competing methods on eight tasks.
%To show the contribution of our collected training data, we retrain several representative baselines on our dataset while keeping the other training settings the same as in their original papers. 
First, we can see that most of the models retrained on our dataset improve over their original counterparts, showing the effectiveness of our training data, which consist of more samples with diverse content. %A few results are slightly worse on some public benchmarks, likely due to domain shift. 
%Since our training data are larger and more diverse, this setting also provides a broader test of each model's native capability and benefits real-world generalization, as shown in the \textbf{Appendix}.
Second, the two variants of PixRestore show the best overall performance, with its variants ranking first or second on most metrics including reference-based metrics and DR-Score. %The main exceptions are deblur, dehaze, and SR, where the reference-based metrics and the DR-Score tend to favor different methods. These degradations typically involve more severe corruption and greater information loss, so methods with slightly higher DR-Scores tend to rely more on aggressive generation.
The results also expose clear differences among methods. The regression model PromptIR remains competitive on denoise, but its performance drops on deblur, rain, haze, low-light, and SR, where the information is lost more severely. Restoration-native diffusion models (DiffUIR, DA-CLIP, FoundIR) generally improve perceptual quality and degradation removal on several tasks. DA-CLIP is relatively strong on de-rainstreak, de-raindrop, and desnow, while FoundIR performs better on dehaze and denoise. This suggests that such models benefit from generative modeling, but their performance varies noticeably across degradations.

Pretrained latent T2I methods show a different trade-off. The SDXL-based FoundIR-v2 attains high DR-Scores but lower fidelity and perceptual performance, which is consistent with the input-detail loss introduced by latent VAE compression. FLUX-based FAPE-IR is among the strongest methods on several degradations (\textit{e.g.}, de-rainstreak and denoise), whereas Flux-IR shows much less consistent performance. This suggests that a stronger generative prior alone does not guarantee better UIR performance. FAPE-IR requires a complex auxiliary design (\textit{e.g.}, Qwen2-VL \cite{qwenvl} and SigLIP in FAPE-IR) to adapt the T2I prior to UIR, as reflected by its substantially large parameter count.

PixRestore is more closely aligned with the restoration objective. Flow matching on patchified pixels preserves spatial evidence that a VAE may weaken, while the hierarchical DINO guidance adapts the conditioning to each degradation rather than relying on a single global prior. Fig.~\ref{fig:syncom} shows visual comparisons. On the desnow and low-light enhancement benchmarks, competing methods often leave residual degradations or recover less faithful details, whereas PixRestore removes the degradations more thoroughly and preserves sharper structures. These gains are achieved with substantially lower inference cost, as shown in Table~\ref{tab:complexity}.

\subsection{Model Complexity Comparisons} 
Table \ref{tab:complexity} compares model size, computation, and latency under the same resolution and hardware. With only 53.7M parameters (including the frozen DINO encoder), 658G FLOPs, and 44\,ms per image, PixRestore is much lighter than other UIR models, running about 7--23\(\times\) faster than PromptIR, DiffUIR, FoundIR, and FAPE-IR.  Its computation is only about 1/181 of FoundIR-v2 and 1/1209 of Flux-IR. In addition, PixRestore-B further improves UIR performance while still keeping the model compact and efficient, with 210.89M parameters, 1842G FLOPs, and 79\,ms latency.
Our results suggest that strong UIR models may not require a lossy latent VAE or a massive T2I prior.

\begin{table*}
  \centering
  \caption{Ablation studies of PixRestore. ``Avg.'' denotes uniform averaging of selected DINO features for conditioning or supervision, while ``Adap.'' denotes adaptive averaging of selected DINO features for conditioning or supervision. ``NFE'' denotes the number of evaluations.}
  \label{tab:ablation_main}
  \setlength{\tabcolsep}{5pt}
  \renewcommand{\arraystretch}{1.10}
  \scriptsize
  \begin{tabular}{c l c c c c c c c}
    \toprule
    ID & Variant & NFE &  Conditioning & Supervision & PSNR$\uparrow$ & SSIM$\uparrow$ & LPIPS$\downarrow$ & MUSIQ$\uparrow$ \\
    \midrule
    A0 & Pixel DiT-S & 10 & None & None & 26.62 & 0.8454 & 0.1593 & 54.32 \\
    A1 & + single-layer conditioning  & 10 & layer 2 & None & 27.07 & 0.8457 & 0.1561 &54.45 \\
    A2 & + single-layer conditioning  & 10 & layer 5 & None & 27.36 & 0.8487 & 0.1489 & 54.64 \\
    A3 & + single-layer conditioning  & 10 & layer 11 & None & 27.12 & 0.8491 & 0.1508 & 54.60 \\
    A4 & + multi-layer conditioning & 10 & Avg. 2 layers & None & 27.62 & 0.8540 & 0.1412 & 54.90 \\
    A5 & + multi-layer conditioning & 10 & Avg. 6 layers & None & 27.72 & 0.8536 & 0.1407 & 55.01 \\
    A6 & + hierarchical loss & 10 & Avg. 6 layers  & Avg. 6 layers & 27.36 & 0.8444 & 0.1239 & 54.59 \\   
    A7 & + adaptive hierarchical visual guidance & 10 & Adap. 6 layers& Adap. 6 layers  & 27.66 & 0.8500 & 0.1209 & 54.97 \\
    \rowcolor{lotuspink}
    A8 & + adaptive hierarchical visual guidance & 4 & Adap. 6 layers& Adap. 6 layers  & 27.75 & 0.8584 & 0.1228 & 54.07 \\
    \rowcolor{lotuspink}
    A9 & + adaptive hierarchical visual guidance & 1 & Adap. 6 layers& Adap. 6 layers  &28.07& 0.8640 & 0.1202  & 53.34 \\
    \rowcolor{lotuspink}
    A10 & + single-step finetuning (PixRestore) & 1 & Adap. 6 layers  & Adap. 6 layers & 28.49 & 0.8589 & 0.1120 & 55.52 \\
    \bottomrule
  \end{tabular}
\end{table*}

\begin{table*}[t]
  \centering
  \caption{Comparison between diffusion pretraining and finetuning with regression training under the same objective and total training iterations.}
  \label{tab:ablation_pretrain}
  \setlength{\tabcolsep}{6pt}
  \renewcommand{\arraystretch}{1.10}
  \scriptsize
  \begin{tabular}{lccccc}
    \toprule
    Scheme & Training iterations& PSNR$\uparrow$ & SSIM$\uparrow$ & LPIPS$\downarrow$ & MUSIQ$\uparrow$ \\
    \midrule
    Regression Training & 350k & 27.00 & 0.8179 & 0.1494 & 52.00 \\
    \rowcolor{lotuspink}
    Flow pretraining + one-step finetuning (ours) & 250k + 100k  & 28.49 & 0.8589 & 0.1120 & 55.52 \\
    \bottomrule
  \end{tabular}
\end{table*}

\subsection{Ablation Studies}

We conduct ablations to verify the main designs of PixRestore. The results are reported in Table~\ref{tab:ablation_main} and Table~\ref{tab:ablation_pretrain}, which are averaged over 15 public benchmarks covering 8 degradation types. All variants use LightningDiT-S in the pixel space as the baseline.

\noindent\textbf{Effect of DINO Conditioning.}
Starting from the plain Pixel DiT-S baseline (A0), adding a single DINO feature consistently improves all metrics. The best single-layer choice is layer 5 (A2), which improves PSNR from 26.62 to 27.36, SSIM from 0.8454 to 0.8487, LPIPS from 0.1593 to 0.1489, and MUSIQ from 54.32 to 54.64. This shows that DINO features provide effective guidance for UIR.

%这里会给个可视化，证明不同layerguide适合于不同的degradation。

% Among the three choices, the middle layer gives the best result, showing that DINO features provide useful restoration guidance. However, different layers perform differently, suggesting that no single layer is sufficient for all degradations.

\noindent\textbf{Single-layer or Multi-layer Guidance.}
Using multiple DINO layers is better than using one fixed layer. Compared with the best single-layer setting A2, averaging 2 layers (A4) improves PSNR from 27.36 to 27.62 and LPIPS from 0.1489 to 0.1412. Averaging 6 layers (A5) further raises PSNR to 27.72 and MUSIQ to 55.01. This shows that shallow and deep DINO layers provide complementary cues for UIR.

\noindent\textbf{Effect of Hierarchical Supervision.}
After introducing the hierarchical feature loss, LPIPS improves clearly from 0.1407 (A5) to 0.1239 (A6), showing better perceptual restoration. However, PSNR drops from 27.72 to 27.36 and SSIM drops from 0.8536 to 0.8444. This suggests that uniform feature-space supervision helps recover more realistic details, but does not give the best overall balance.

\noindent\textbf{Adaptive Hierarchical Visual Guidance.}
We further compare A6 and A7. Both of them use multi-layer conditioning and supervision, but A6 uses uniform averaging while A7 uses adaptive layer weighting. A7 improves PSNR from 27.36 to 27.66, SSIM from 0.8444 to 0.8500,  MUSIQ from 54.59 to 54.97, and LPIPS from 0.1239 to 0.1209. This shows that adaptive guidance better exploits the degradation-dependent reliability of DINO layers.

\noindent\textbf{Different NFE.}
We further study the effect of reducing the number of function evaluations (NFE). Compared with A7 which uses 10 NFE, A8 with 4 NFE improves PSNR from 27.66 to 27.75 and SSIM from 0.8500 to 0.8584, while LPIPS changes from 0.1209 to 0.1228. When NFE is further reduced to 1, A9 still improves PSNR to 28.07 and SSIM to 0.8640, with LPIPS of 0.1202. MUSIQ drops from 54.97 to 54.07 and 53.34, but the overall results remain competitive. Interestingly, reducing NFE slightly improves PSNR/SSIM in our setting, possibly because fewer Euler updates reduce the accumulation of integration errors and over-smoothing. These results suggest that PixRestore maintains strong restoration quality even in the one-step setting.

\noindent\textbf{Effect of Single-step Finetuning.}
We finetune the multi-step model into a one-step generator. Compared with A9, the final PixRestore (A10) improves PSNR from 28.07 to 28.49, LPIPS from 0.1202 to 0.1120, and MUSIQ from 53.34 to 55.52, while keeping NFE at 1. This shows that one-step fine-tuning improves restoration quality while retaining one-step efficiency.

\noindent\textbf{Flow Pretraining vs. Regression Training.}
Table~\ref{tab:ablation_pretrain} compares our training pipeline with direct regression training under the same backbone, loss functions (with one-step finetuning), and total training iterations. Flow pretraining followed by one-step finetuning improves PSNR from 27.00 to 28.49, SSIM from 0.8179 to 0.8589, LPIPS from 0.1494 to 0.1120, and MUSIQ from 52.00 to 55.52. This shows that the gain comes from the flow-based pretraining stage.

\begin{figure}[t]
  \centering
  \includegraphics[width=0.65\linewidth]{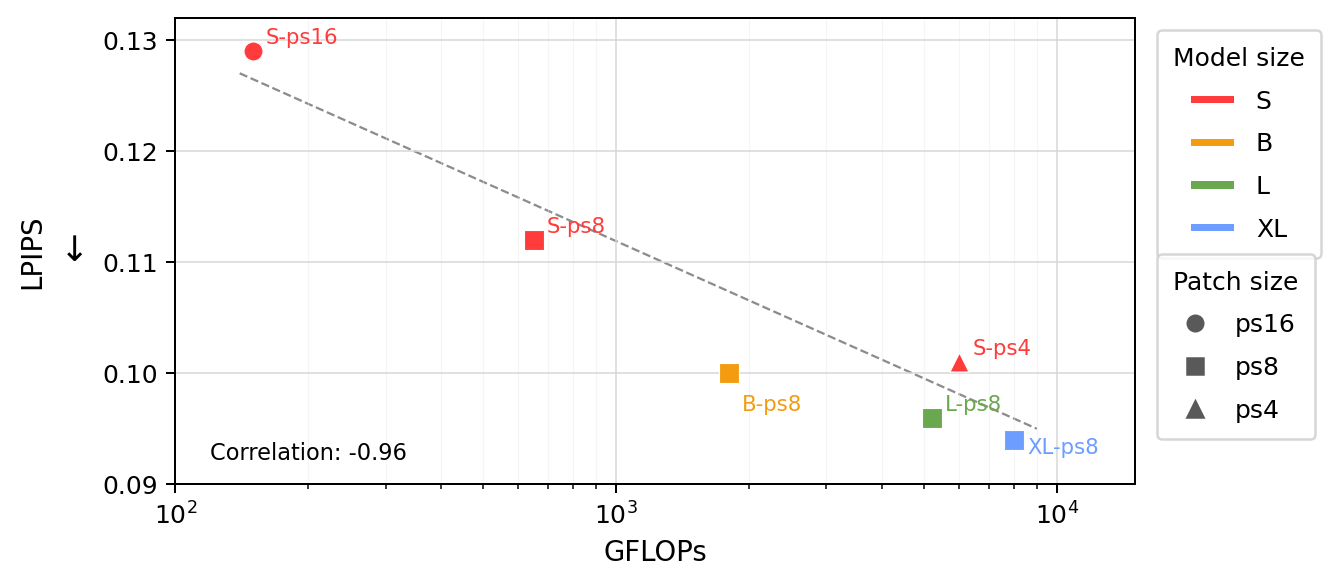}
  % \vspace{-3mm}
  \caption{Scaling behavior of PixRestore under varying backbone GFLOPs and patch sizes.}
  \label{fig:scale}
  % \vspace{-4mm}
\end{figure}

\subsection{Scalability}
Fig.~\ref{fig:scale} illustrates the scalability of PixRestore by varying the Transformer size and pixel patch size. A smaller patch yields more image tokens and thus more computation, which improves LPIPS for the same backbone; for example, the S model improves the LPIPS from about 0.129 with \(p=16\) to about 0.101 with \(p=4\). Enlarging the backbone at a fixed patch size brings a similar gain. Across all configurations, LPIPS decreases as GFLOPs increase, with a correlation of $-0.96$. Consistent with the scaling behavior reported for DiT \cite{dit}, this smooth trend suggests that PixRestore can benefit from scaling along two axes: a smaller patch preserves more local evidence, while a larger backbone provides stronger global modeling.

\begin{table*}[t]
    \caption{Quantitative comparison on real-world test set. The best and second-best results for each metric are highlighted in \textcolor{red}{\textbf{red bold}} and \textcolor{blue}{\textit{blue italic}}, respectively. Retrained methods are marked with $^*$. `PR' denotes the proposed PixRestore, the results of which are shaded in pink.} %We report two variants of our method with different model sizes.}
    \label{tab:real_dr_score_comparison}
    \centering
    \small
    \setlength{\tabcolsep}{2.2pt}
    \renewcommand{\arraystretch}{1.08}
    \setlength{\aboverulesep}{0.2pt}
    \setlength{\belowrulesep}{0.2pt}
    \resizebox{\textwidth}{!}{%
    \begin{tabular}{l l c c c c c c c c c c c c c c >{\columncolor{lotuspink}}c >{\columncolor{lotuspink}}c}
      \toprule
      Degradation & Metric & PromptIR & PromptIR$^*$ & DiffUIR & DA-CLIP & DA-CLIP$^*$ & FoundIR & FoundIR$^*$ & UniRestore & FoundIR-v2 & FoundIR-v2$^*$ & Flux-IR & Flux-IR$^*$ & FAPE-IR & FAPE-IR$^*$ & PR & PR-B \\
      \midrule
      \multirow{3}{*}{De-rainstreak}
        & MUSIQ$\uparrow$ & 59.90 & 60.06 & 61.02 & 62.23 & 59.94 & 61.00 & 60.87 & 62.58 & \textcolor{red}{\textbf{63.06}} & 62.57 & \textcolor{blue}{\textit{62.98}} & 60.23 & 61.07 & 59.53 & 62.62 & 62.78 \\
        & Affine-NR$\downarrow$ & -0.90 & -0.90 & -0.92 & -0.91 & -0.91 & -0.90 & -0.92 & -0.89 & -0.96 & -0.96 & -0.94 & -0.89 & -1.00 & -1.00 & \textcolor{blue}{\textit{-1.02}} & \textcolor{red}{\textbf{-1.03}} \\
        & DR-Score $\uparrow$ & 30.85 & 30.05 & 33.77 & 33.62 & 51.68 & 31.48 & 38.83 & 32.87 & 49.07 & 71.72 & 25.50 & 28.37 & 64.73 & \textcolor{red}{\textbf{76.32 }}& 72.12 & \textcolor{blue}{\textbf{74.69}} \\
      \midrule
      \multirow{3}{*}{Deblur}
        & MUSIQ$\uparrow$ & 34.28 & 32.57 & 41.64 & 46.61 & 43.92 & 32.85 & 34.10 & 49.26 & \textcolor{blue}{\textit{69.99}} & \textcolor{red}{\textbf{72.78}} & 55.26 & 65.73 & 44.89 & 45.84 & 53.37 & 56.56 \\
        & Affine-NR$\downarrow$ & -0.73 & -0.70 & -0.78 & -0.77 & -0.76 & -0.70 & -0.72 & -0.81 & \textcolor{blue}{\textit{-1.04}} & \textcolor{red}{\textbf{-1.10}} & -0.87 & -1.00 & -0.80 & -0.83 & -0.88 & -0.93 \\
        & DR-Score $\uparrow$ & 26.56 & 31.08 & 28.12 & 48.78 & 45.22 & 26.58 & 29.35 & 46.77 & 66.66 & \textcolor{red}{\textbf{74.94}} & 38.09 & 72.18 & 63.35 & 65.41 & 65.43 & \textcolor{blue}{\textit{73.08}} \\
      \midrule
      \multirow{3}{*}{De-raindrop}
        & MUSIQ$\uparrow$ & 64.18 & 63.47 & 64.04 & \textcolor{blue}{\textit{66.26}} & 54.46 & 61.85 & 60.62 & 63.87 & 65.81 & 57.74 & 65.60 & \textcolor{red}{\textbf{66.28}} & 52.71 & 39.61 & 47.99 & 54.54 \\
        & Affine-NR$\downarrow$ & -0.83 & -0.78 & -0.82 & -0.87 & -0.71 & -0.77 & -0.71 & -0.78 & -0.83 & -0.79 & \textcolor{blue}{\textit{-0.92}} & \textcolor{red}{\textbf{-1.02}} & -0.85 & -0.76 & -0.72 & -0.80 \\
        & DR-Score$\uparrow$ & 25.70 & 26.15 & 22.10 & 37.63 & 44.65 & 26.57 & 32.38 & 26.00 & 34.30 & 62.59 & 36.05 & 36.03 & \textcolor{blue}{\textit{80.73}} & \textcolor{red}{\textbf{80.82}} & 74.88 & 80.62 \\
      \midrule
      \multirow{3}{*}{Desnow}
        & MUSIQ$\uparrow$ & 58.96 & 59.27 & 59.92 & 59.74 & 59.36 & 60.24 & 60.15 & 60.30 & \textcolor{blue}{\textit{63.35}} & \textcolor{red}{\textbf{63.46}} & 58.33 & 62.12 & 57.96 & 58.68 & 62.69 & 62.65 \\
        & Affine-NR$\downarrow$ & -0.74 & -0.75 & -0.77 & -0.78 & -0.77 & -0.75 & -0.75 & -0.73 & -0.83 & -0.86 & -0.76 & -0.89 & -0.86 & -0.88 & \textcolor{blue}{\textit{-0.90}} & \textcolor{red}{\textbf{-0.91}} \\
        & DR-Score $\uparrow$ & 26.90 & 34.17 & 39.53 & 45.99 & 46.69 & 28.01 & 32.45 & 32.13 & 46.90 & 64.89 & 38.63 & 50.63 & 71.57 & 71.73 & \textcolor{blue}{\textit{71.86}} & \textcolor{red}{\textbf{74.70}} \\
      \midrule
      \multirow{3}{*}{Dehaze}
        & MUSIQ$\uparrow$ & 59.83 & 60.40 & 59.66 & 61.00 & 60.21 & 60.26 & 60.55 & 60.85 & \textcolor{blue}{\textit{63.38}} & 61.32 & \textcolor{red}{\textbf{63.39}} & 60.32 & 59.77 & 60.33 & 61.93 & 61.47 \\
        & Affine-NR$\downarrow$ & -0.88 & -0.88 & -0.87 & -0.88 & -0.87 & -0.87 & -0.88 & -0.82 & -0.90 & -0.86 & -0.89 & -0.83 & -0.88 & -0.90 & \textcolor{red}{\textbf{-0.93}} & \textcolor{blue}{\textit{-0.93}} \\
        & DR-Score $\uparrow$ & 32.90 & 37.68 & 24.55 & 32.36 & 28.86 & 31.02 & 32.43 & \textcolor{red}{\textbf{47.43}} & 42.61 & 37.52 & 40.37 & 40.74 & 32.67 & 35.94 & \textcolor{blue}{\textit{44.51}} & 42.06 \\
      \midrule
      \multirow{3}{*}{Low-light}
        & MUSIQ$\uparrow$ & 47.67 & 55.03 & 54.21 & \textcolor{red}{\textbf{64.66}} & 53.18 & 49.61 & 57.10 & 48.47 & \textcolor{blue}{\textit{63.33}} & 61.31 & 54.10 & 52.75 & 49.91 & 57.60 & 58.17 & 58.63 \\
        & Affine-NR$\downarrow$ & -0.88 & -0.92 & -0.71 & -0.95 & -0.90 & -0.89 & -0.98 & -0.85 & \textcolor{red}{\textbf{-1.00}} & -0.94 & -0.93 & -0.91 & -0.89 & \textcolor{blue}{\textit{-0.99}} & -0.98 & \textcolor{blue}{\textit{-0.99}} \\
        & DR-Score$\uparrow$ & 31.65 & 67.03 & 59.33 & 66.24 & 65.10 & 31.70 & 65.05 & 38.87 & \textcolor{blue}{\textit{71.25}} & 66.62 & 58.07 & 54.71 & 38.70 & \textcolor{red}{\textbf{75.09}} & 62.38 & 62.13 \\
      \midrule
      \multirow{3}{*}{Average}
        & MUSIQ$\uparrow$ & 54.14 & 55.13 & 56.75 & 60.08 & 55.18 & 54.30 & 55.56 & 57.55 & \textcolor{red}{\textbf{64.82}} & \textcolor{blue}{\textit{63.20}} & 59.45 & 61.24 & 54.39 & 53.60 & 57.80 & 59.44 \\
        & Affine-NR$\downarrow$ & -0.83 & -0.82 & -0.81 & -0.86 & -0.82 & -0.81 & -0.83 & -0.81 & \textcolor{red}{\textbf{-0.93}} & -0.92 & -0.89 & -0.92 & -0.88 & -0.88 & -0.91 & \textcolor{red}{\textbf{-0.93}} \\
        & DR-Score $\uparrow$ & 29.09 & 37.69 & 34.57 & 44.10 & 47.03 & 29.23 & 38.42 & 37.51 & 51.80 & 63.05 & 39.45 & 47.11 & 58.63 & \textcolor{blue}{\textit{67.55}} & 65.20 & \textcolor{red}{\textbf{67.88}} \\
      \bottomrule
    \end{tabular}%
    }
  \end{table*}

\begin{figure*}[h]
  \centering
  \includegraphics[width=0.9\linewidth]{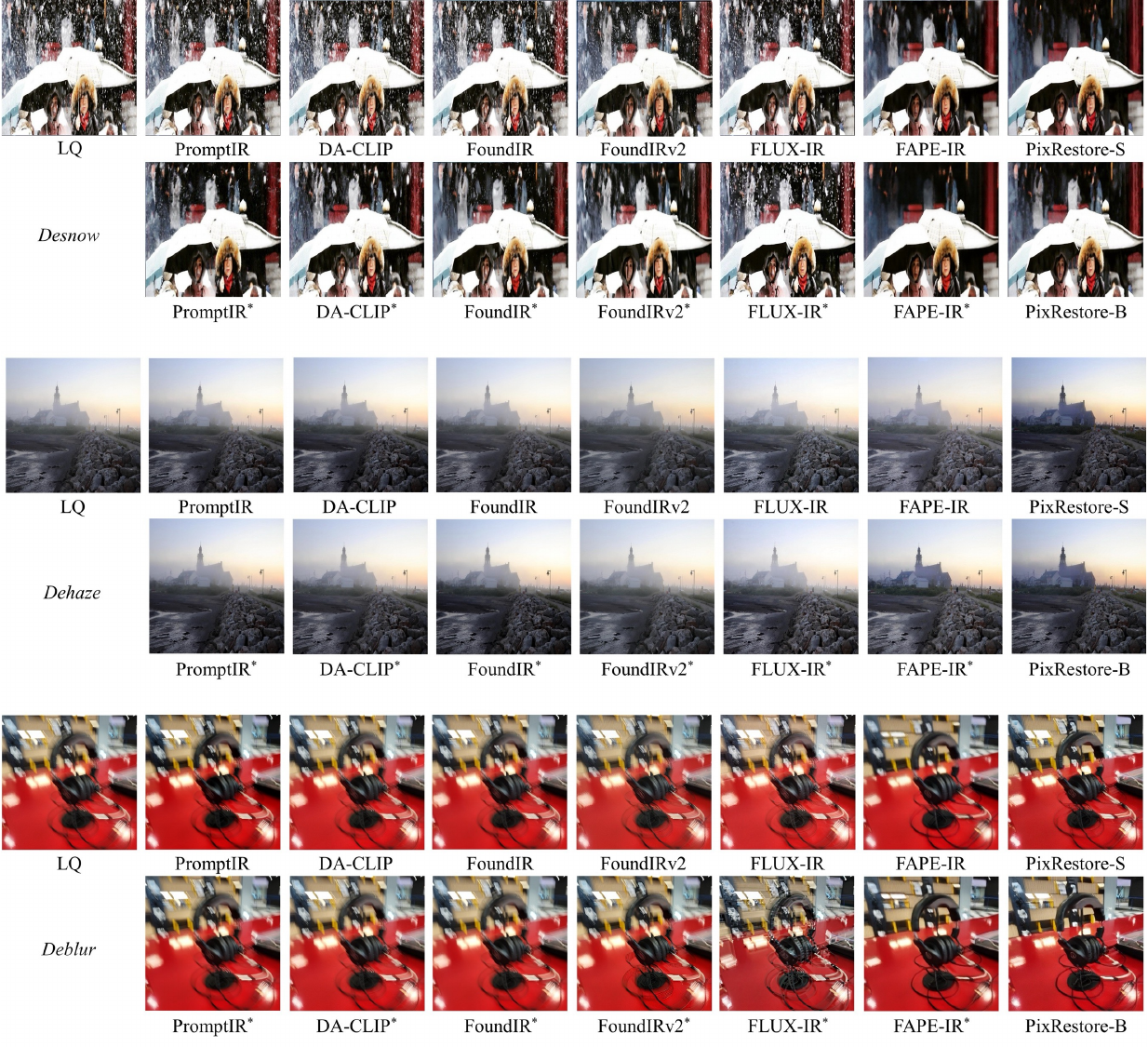}
  % \vspace{-4mm}
  \caption{Visual comparisons on real-world desnow, dehaze, and deblur cases. PixRestore can remove degradations more effectively and recover cleaner details and sharper structures with fewer artifacts.}
  \label{fig:supp-real1}
  % \vspace{-5mm}
\end{figure*}

\begin{figure*}[h]
  \centering
  \includegraphics[width=0.9\linewidth]{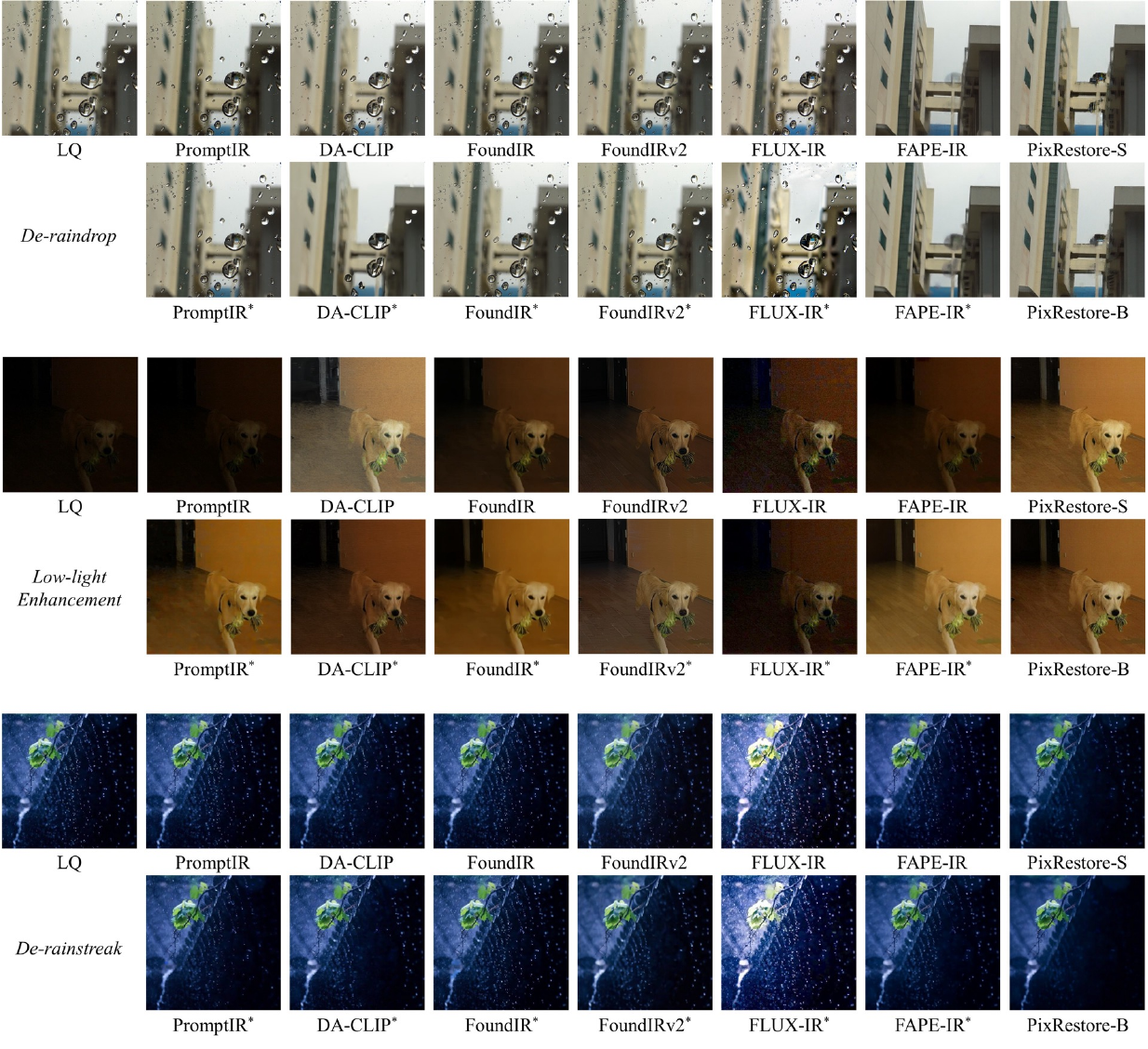}
  % \vspace{-4mm}
  \caption{Visual comparisons on real-world de-raindrop, low-light enhancement, and de-rainstreak cases. PixRestore can remove degradations more effectively and recover cleaner details and sharper structures with fewer artifacts.}
  \label{fig:supp-real2}
  % \vspace{-5mm}
\end{figure*}

\subsection{Generalization to Real-world Test Set}
We test all methods on real-world test data to evaluate their generalization ability to out-of-domain scenarios. Since no GT images are available, we report MUSIQ and AFINE-NR as auxiliary no-reference quality metrics, while relying primarily on DR-Score and visual comparisons to assess degradation removal. The results are shown in Table~\ref{tab:real_dr_score_comparison}. We see that retraining on our training data improves the average DR-Score of many existing methods, but the improvements are not consistent across degradation types. For example, retrained DA-CLIP substantially improves its DR-Score on de-rainstreak, from 33.62 to 51.68, but its score decreases on deblur, from 48.78 to 45.22. This indicates that gains on one degradation may come at the expense of performance drop on another under the unified restoration setting.

It can also be seen that some strong competing methods show leading scores of MUSIQ and Afine-NR on some degradation types. For example, FoundIR-v2 and FoundIR-v2$^{*}$ obtain the highest average MUSIQ scores,  but perform poorly on raindrop removal.
Flux-IR$^{*}$ performs the best on MUSIQ for de-raindrop, but its degradation removal performance on rainstreak is poor. 
As we discussed in Sec. \ref{exp-set}, the NR-IQA metrics such as MUSIQ and Afine-NR cannot reflect the real degradation removal performance, and this is why we propose DR-Score to more reliably measure this important ability. From  Table~\ref{tab:real_dr_score_comparison}, we see that 
PixRestore-B achieves the best average DR-Score of 67.88, followed by FAPE-IR$^{*}$ at 67.55. This suggests that PixRestore offers a better overall balance between degradation removal and perceptual quality.

Figs.~\ref{fig:supp-real1} and \ref{fig:supp-real2} present visual comparisons on six real-world degradation types. We see that the original versions of many baseline methods often fail to remove the target degradation sufficiently, while their retrained versions on our training data shown better degradation removal performance, yet they still suffer from residual artifacts, over-smoothing, color bias, or unstable detail reconstruction. In contrast, PixRestore demonstrates more balanced restoration results, achieving stronger degradation removal together with more natural color, clearer structures, and fewer artifacts. For example, in the desnow case, some competing methods leave visible snow residues, whereas PixRestore restores a cleaner image with better structural clarity. In the dehaze example, many competing methods  produce grayish or flat-looking outputs with limited visibility, while PixRestore reveals clearer scene content and more natural contrast.

\section{Conclusion}
We presented PixRestore, a VAE-free pixel-space diffusion transformer for unified image restoration. By performing flow matching directly on patchified pixels and incorporating adaptive hierarchical DINO guidance in both conditioning and supervision, PixRestore achieved faithful restoration with strong perceptual quality across diverse degradations while remaining compact and efficient with single-step inference. Larger PixRestore variants further improve performance, demonstrating the scalability of our design.

\noindent\textbf{Limitations.} PixRestore has several limitations. First, due to the DiT architecture and fixed tokenization setting, a model trained at one resolution cannot be directly extended to higher-resolution without retraining or architectural modification. Second, compared with billion-scale T2I models, the compact PixRestore backbone may encounter difficulties under extremely information-scarce degradations. Finally, our DR-Score relies on a proprietary VLM, introducing potential differences across model updates. In future work, we will explore more resolution-flexible architectures, stronger visual priors, and dedicated IR-specific quality metrics.

%==========================================================================
% Bibliography
%==========================================================================
{\small
\bibliography{ref}
}

\appendix

\clearpage
\setcounter{table}{0}
\setcounter{equation}{0}
\setcounter{figure}{0}
\renewcommand{\thetable}{S.\arabic{table}}
\renewcommand{\theequation}{S.\arabic{equation}}
\renewcommand{\thefigure}{S.\arabic{figure}}
\begin{center}
    {\LARGE\bfseries Appendix}
\end{center}

\noindent The following materials are provided in this appendix:
\begin{itemize}
    \item Details of training data and real-world test data collection (see Sec. 4.1 of the main paper).
    \item Pixel-space vs. latent-space diffusion models for unified image restoration (see Sec. 3.1 of the main paper).
    \item Visual foundation prior for unified image restoration (see Sec. 3.2 of the main paper).
    \item Details of DR-Score (see Sec. 4.1 of the main paper).
    \item More public benchmark comparisons, including the per-dataset numerical comparisons and more visual comparisons (Sec. 4.2 of the main paper).
    
\end{itemize}

\section{Training Data and Real-World Test Data}
We build a training corpus of about 2.83M images spanning eight restoration tasks: deblur, dehaze, denoise, de-rainstreak, de-raindrop, desnowing, low-light enhancement, and super-resolution (SR). During training, samples are drawn from these tasks with equal probability. Table~\ref{tab:training-data-sources} summarizes the training data sources for each degradation type, highlighting the diversity of both degradation patterns and scene content.

For haze, rainstreak, and snow, which depend strongly on scene depth, we further incorporate commonly used datasets with image-depth pairs and synthesize degradations following the pipelines of MioIR \cite{kong2024towards} and RealRestorer \cite{yang2026realrestorer}. For denoising, we synthesize noisy inputs by adding Gaussian noise with three noise levels, \textit{i.e.}, $\sigma=15, 25,$ and $50$, following DnCNN \cite{zhang2017beyond}. For SR, we evaluate the $4\times$ setting, with low-quality (LQ) inputs resized to $512\times512$ to match the high-quality (HQ) images before feeding them into the model. There is no image-level overlap between the training data and the evaluated public benchmarks. For datasets without an official train/test split, such as PolyU \cite{polyunoise} and ScreenSR \cite{vosr}, we reserve a portion of about 10\% for testing and use the rest for training.

For real-world evaluation, we collect benchmarks covering six degradation types, each containing 100 real photographs, as summarized in Table~\ref{tab:real-test-data-sources}. These images are used to evaluate the generalization ability of restoration methods and do not have ground-truth (GT) references. We do not include real-world denoising and SR in this benchmark because they are difficult to define as isolated degradations in real images. In practice, real-world low-quality (LQ) images usually contain mixed degradations: low-light images are often accompanied by noticeable noise, while real-world low-resolution images are also commonly affected by blur and noise. Therefore, we focus on six representative real-world degradation types, which are sufficient to assess the generalization performance of different methods. All real-world images are resized to $512\times512$ for testing.

\begin{table*}[htbp]
\centering
\caption{Training data sources for each degradation type.}
\label{tab:training-data-sources}
\small
\setlength{\tabcolsep}{6pt}
\begin{tabular}{p{3.0cm}p{12.8cm}}
\toprule
\textbf{Degradation} & \textbf{Training data} \\
\midrule
Deblur
& GoPro \cite{GoPro}, RealBlur \cite{rim2020realblur}, UHD-Blur \cite{uhdblurhaze}, LSD-Defocus \cite{lsddblur} \\
\addlinespace
Dehaze
& RESIDE \cite{RESIDE}, UHD-Haze \cite{uhdblurhaze}, WeatherBench-haze \cite{guan2025weatherbench},
  UniSer-Haze \cite{zhang2025uniser}, synthetic data from UrbanSyn \cite{urbansyn}, BlendedMVS \cite{yao2020blendedmvs} and MegaDepth \cite{li2018megadepth}\\
\addlinespace
De-raindrop
& RaindropClarity \cite{jin2024raindropclarity}, RainDS-Real-RainDrop \cite{rainds}, \\
\addlinespace
de-rainstreak
& Rain13K \cite{zamir2022restormer}, RealRain-1k \cite{realrain}, UAV-Rain1k \cite{UAV-Rain1k}, FoundIR-rain \cite{li2025foundir},
  RainDS-Real-RainStreak \cite{rainds}, synthetic data from UrbanSyn \cite{urbansyn}, \\
\addlinespace
Desnow
& Snow100K \cite{snow100k}, WeatherBench-snow \cite{guan2025weatherbench}, synthetic data from UrbanSyn \cite{urbansyn}, \\
\addlinespace
Denoise
& SIDD \cite{sidd}, PolyU \cite{polyunoise}, synthetic Gaussian noise DF2K \cite{div2k, flickr}\\
\addlinespace
Low-light \\ enhancement
& LOL \cite{LoL}, UHD-LL \cite{uhdll}, DarkFace \cite{darkface2019}, FoundIR-low-light \cite{li2025foundir},
  NTIRE-LLIE \cite{liu2024ntire} \\
\addlinespace
Super-resolution (SR)
& RealESRGAN degradation from DF2K \cite{div2k,flickr}, RealSR \cite{realsr}, ScreenSR \cite{vosr} \\
\bottomrule
\end{tabular}
\end{table*}\begin{table*}[t]
\centering
\caption{Real-world test data sources for each degradation type.}
\label{tab:real-test-data-sources}
\small
\setlength{\tabcolsep}{6pt}
\begin{tabular}{p{3.0cm}p{12.8cm}}
\toprule
\textbf{Degradation} & \textbf{Real-world test data} \\
\midrule
Deblur
& GyroBlur-Real \cite{yang2024gyro} \\
\addlinespace
Dehaze
& RTTS and OpenReal-fog \cite{jarvisir2025} \\
\addlinespace
De-raindrop
& OpenReal-raindrop \cite{jarvisir2025}  \\
\addlinespace
De-rainstreak
& OpenReal-rainstreak \cite{jarvisir2025} and
  DiffUIR\cite{diffuir} \\
\addlinespace
Desnow
& Snow100K-realistic \cite{snow100k} and OpenReal-snow \cite{jarvisir2025}  \\
\addlinespace
\addlinespace
Low-light \\ enhancement
& OpenReal-night \cite{jarvisir2025} and ExDark \cite{Exdark} \\
\addlinespace
\bottomrule
\end{tabular}
\end{table*}

\begin{table*}
\centering
\scriptsize
\caption{Pixel-space and latent-space comparison on 8 degradation types. We report comparisons on PSNR$\uparrow$/LPIPS$\downarrow$/MUSIQ$\uparrow$.}
\label{supp_tab:pixel_vs_latent}
\setlength{\tabcolsep}{5pt}

\begin{tabular}{lcccc}
\toprule
\textbf{Degradation}
& \textbf{Latent DiT with FLUX-VAE}
& \textbf{Latent DiT with Qwen-VAE}
& \textbf{Latent DiT with SD2-VAE}
& \textbf{Pixel DiT} \\
\midrule
SR
& 25.59/0.2038/58.68
& 25.62/\textbf{0.1992}/\textbf{62.25}
& 25.03/0.2586/58.53
& \textbf{26.48}/0.2067/57.80 \\
Deblur
& 26.66/0.1617/46.53
& 26.93/\textbf{0.1545}/\textbf{46.69}
& 26.00/0.1921/45.23
& \textbf{27.17}/0.1850/41.76 \\
Dehaze
& 15.31/0.2253/54.56
& 15.16/0.2391/52.50
& 15.08/0.2659/53.31
& \textbf{24.78}/\textbf{0.1020}/\textbf{58.17} \\
Denoise
& 31.97/\textbf{0.0952}/47.22
& 32.33/0.1021/48.37
& 30.65/0.1252/47.90
& \textbf{32.58}/0.1179/\textbf{51.32} \\
De-raindrop
& 20.43/0.2257/62.79
& 20.63/0.2463/64.71
& 20.01/0.2751/62.46
& \textbf{23.19}/\textbf{0.1707}/\textbf{66.66} \\
De-rainstreak
& 25.07/0.1478/46.49
& 25.34/0.1821/47.63
& 24.53/0.1876/46.00
& \textbf{30.19}/\textbf{0.1350}/\textbf{49.43} \\
Desnow
& 27.82/0.1304/49.00
& 28.41/\textbf{0.1190}/49.39
& 27.30/0.1483/\textbf{49.45}
& \textbf{28.96}/0.1301/48.58 \\
Low-light enhancement
& 10.82/0.4572/40.64
& 10.82/0.4530/42.13
& 10.81/0.4834/39.73
& \textbf{20.80}/\textbf{0.2122}/\textbf{57.98} \\
\midrule
\textbf{Overall}
& 22.63/0.2109/50.86
& 22.80/0.2181/51.87
& 22.10/0.2483/50.38
& \textbf{26.62}/\textbf{0.1593}/\textbf{54.32} \\
\bottomrule
\end{tabular}
\end{table*}

\section{Pixel-space vs. Latent-space}
In the main paper, we compare pixel-space and latent-space diffusion models using the average results over all eight degradation types, and pixel diffusion shows a clear advantage for unified image restoration (UIR). Here, we further present the per-degradation results in Table~\ref{supp_tab:pixel_vs_latent}.

We conduct the comparison under the same LightningDiT-S \cite{vavae} backbone and training protocol. The pixel model directly operates on RGB patches with a patch size of 8, while the latent models take latent representations encoded by FLUX-VAE \cite{FLUX}, Qwen-VAE \cite{Qwen-Image-Edit}, and SD2-VAE \cite{sd2.1} with a latent patch size of 1. This design keeps the same spatial compression ratio across models for a fair comparison. None of these models uses an external visual foundation model, such as DINOv2 \cite{dinov2}. During inference, all models use 10 sampling steps and a guidance scale of 1.0. We evaluate them on 15 public benchmarks covering eight degradation types. Following the evaluation protocol in Sec.~4.1 of the main paper, we first compute each metric on each test dataset, and then average the results with equal weight within each degradation type.

More specifically, on SR and deblurring, the latent diffusion model with Qwen-VAE achieves better perceptual scores of LPIPS (0.1992 vs.\ 0.2067 on SR, 0.1545 vs.\ 0.1850 on deblurring) and MUSIQ (62.25 vs.\ 57.80 on SR, 46.69 vs.\ 41.76 on deblurring). For dehazing, de-raindrop removal, de-rainstreak removal, and low-light enhancement, Pixel DiT performs best on all three metrics. The gains are especially large on dehazing (24.78 PSNR, 0.1020 LPIPS, and 58.17 MUSIQ) and low-light enhancement (20.80 PSNR, 0.2122 LPIPS, and 57.98 MUSIQ), far surpassing all latent alternatives. For denoising, Pixel DiT achieves the best PSNR and MUSIQ (32.58 and 51.32), while FLUX-VAE gives the best LPIPS (0.0952). For desnowing, Pixel DiT leads in PSNR (28.96), whereas Qwen-VAE and SD2-VAE lead in LPIPS (0.1190) and MUSIQ (49.45), respectively. Overall, latent models can achieve better perceptual scores on several specific degradations, but pixel-space modeling delivers consistently the best restoration fidelity across diverse tasks.

\section{Visual Foundation Prior} 
We leverage a frozen visual foundation encoder to provide dense visual cues, as the main paper presents.
To choose the encoder, we compare four frozen candidates: CLIP \cite{CLIP}, DINOv2 \cite{dinov2}, MAE \cite{MAE}, and SigLIP \cite{siglip}. All candidates share the same ViT-B architecture and parameter count, and we extract dense tokens from the same layer position (the 11th layer). Meanwhile, the pixel DiT, training data, and optimization schedule are kept fixed, and only the frozen encoder differs, so that the comparison cleanly reflects the effectiveness of different pretrained vision models for UIR. During inference, all models use 10 sampling steps and a guidance scale of 1.0. Table~\ref{tab:visual_prior} reports metrics averaged over 15 public benchmarks covering 8 degradation types.
The self-supervised DINOv2 achieves the best overall fidelity and perceptual balance, while CLIP obtains a marginally higher MUSIQ score. We attribute this to its pretraining objective. Through self-distillation over global and local views, DINO learns dense, spatially precise tokens that preserve the fine structures and textures that restoration must recover, while its tokens remain semantically discriminative and encode high-level content. This dual property is what UIR needs: spatial details are used to reconstruct faithful pixels, and semantic cues are used to distinguish reliable content from degradation. In contrast, CLIP and SigLIP are aligned to text and emphasize global semantics, discarding much of the spatial detail, while MAE targets low-level pixel reconstruction and yields less discriminative structural cues. We therefore adopt DINOv2 as the default encoder.

\begin{table}[t]
  \caption{Comparison of frozen visual foundation priors under the same pixel DiT setting. All encoders share the ViT-B architecture and parameter count. Features of the same layer are selected. Metrics are averaged over 8 degradation types. Best results are highlighted in \textbf{bold}.}
  \centering
  \scriptsize
  \begin{tabular}{lcccc}
    \toprule
    Visual Prior & PSNR (dB)\,$\uparrow$ & SSIM\,$\uparrow$ & LPIPS\,$\downarrow$ & MUSIQ\,$\uparrow$ \\
    \midrule
    CLIP-B       & 26.61 & 0.8441 & 0.1598 & \textbf{54.30} \\
    MAE-B          & 26.70 & 0.8454 & 0.1593 & 54.20 \\
    SigLIP-B     & 26.66 & 0.8449 & 0.1592 & 54.19 \\
    \rowcolor{lotuspink}
    \textbf{DINOv2-B}     & \textbf{27.25} & \textbf{0.8500} & \textbf{0.1531} & 54.21 \\
    \bottomrule
  \end{tabular}
  \label{tab:visual_prior}
\end{table}

\begin{figure*}[t]
  \centering
  \includegraphics[width=0.95\linewidth]{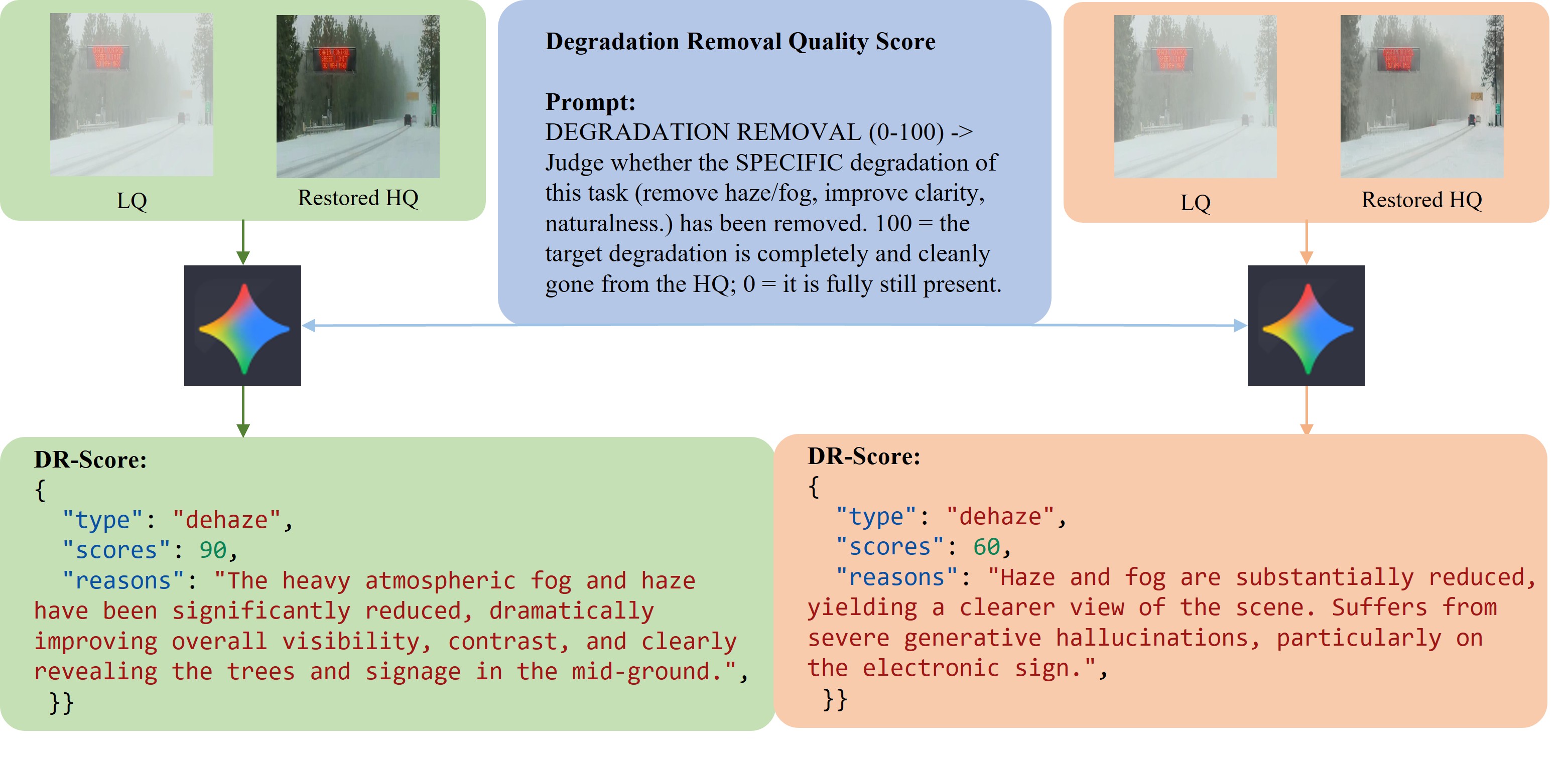}
  \vspace{-4mm}
  \caption{Illustration of DR-Score. Given the LQ input, the restored result, and the task description, the VLM judges whether the target degradation has been removed. A better restoration receives a higher DR-Score.}
  \label{fig:supp-drscore}
  \vspace{-5mm}
\end{figure*}

\begin{figure*}[h]
  \centering
  \includegraphics[width=0.95\linewidth]{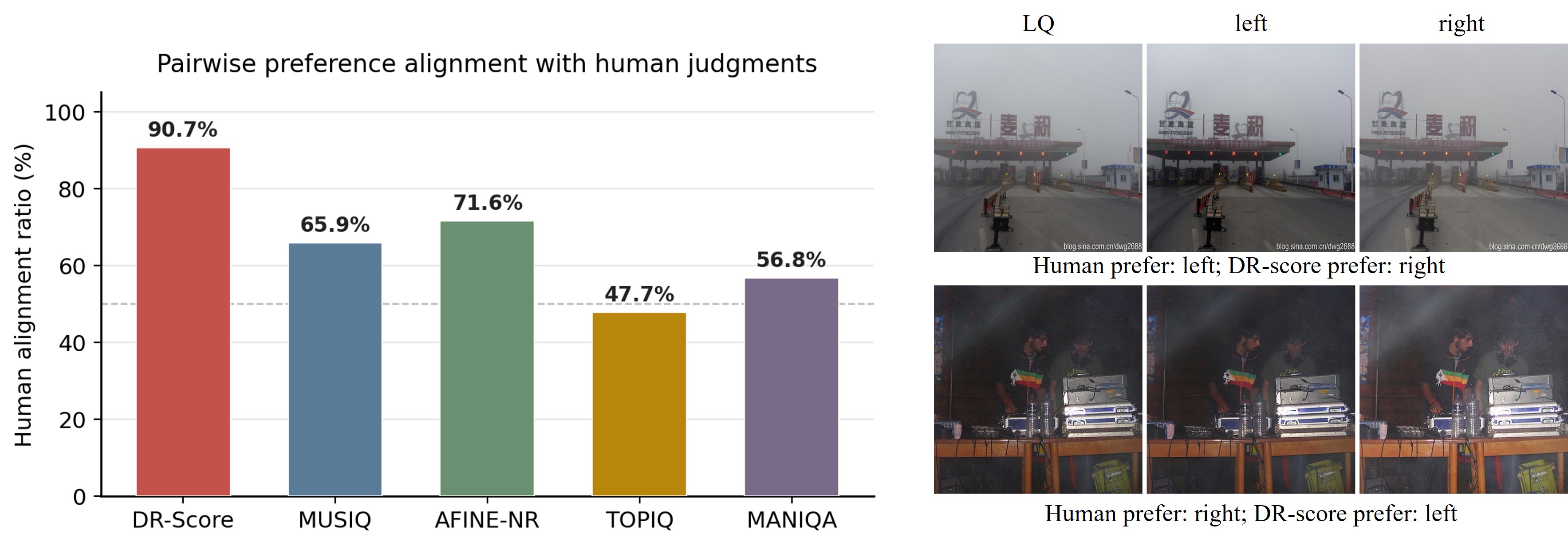}
  \vspace{-4mm}
  \caption{Human alignment analysis of different no-reference metrics with pairwise human preferences. Left: alignment ratios between metric-induced rankings and human judgments across real-world UIR tasks. Right: representative failure cases where two restored results are visually similar with subtle appearance differences but lead to disagreement between DR-Score and human judgment.}
  \label{fig:human-align}
  %\vspace{-5mm}
\end{figure*}
% Among the available baselines, PromptIR remains modest (29.09 average DR-score), reflecting the limited robustness of direct deterministic regression on real scenes. FoundIR-v2 performs strongly on deblur and low-light, while FAPE-IR is competitive on rain-related degradations and achieves the best available average baseline result (58.96). Our method remains competitive overall: Ours reaches 65.96 average DR-score, and Ours-L further improves to 70.41, with especially strong performance on derainstreak, deraindrop, and desnow. Fig. \ref{fig:syncom} shows that our method produces visually faithful results, removing blur and rain artifacts effectively while preserving faces, textures, and colors.

\section{DR-Score Details}

As discussed in the main paper, common full-reference metrics require GT images, which are unavailable for many real-world benchmarks. Existing no-reference metrics mainly assess overall visual quality, but they often fail to capture whether the target degradation has been removed from the restored image. To address this issue, we introduce DR-Score, a vision-language model (VLM)-based auxiliary metric to evaluate degradation removal performance in UIR tasks, rather than to replace standard metrics.

\noindent\textbf{Evaluation Protocol.}
We use Gemini 3.1 pro \cite{gemini3-1pro} as the evaluator, as shown in Fig. \ref{fig:supp-drscore}. For each test sample, we provide the VLM with the low-quality (LQ) input image, the restored output image, and the restoration task type.
Each task type is paired with a short description of the target goal. The VLM compares the restored image with the LQ input and is asked to judge whether the target task has been completed. Specifically, for deblur, the VLM is asked to judge whether the motion or defocus blur has been removed and sharp details are recovered. For dehaze, VLM judges whether the haze or fog has been removed and whether the scene becomes clearer. For de-rainstreak and de-raindrop, it judges whether rain streaks or raindrops are removed and whether the occluded background is restored. For desnow, it judges whether snow is removed. For low-light enhancement, it judges whether the image is properly brightened, denoised, and detailed. For denoising, it judges whether noise is removed while textures are preserved. For SR, it judges whether sharp details are restored without over-smoothing or obvious artifacts.
The score ranges from 0 to 100, with 100 indicating that the target degradation is removed completely and cleanly and 0 indicating that it is still fully present.

We also ask the VLM to return the task type, the score, and a short reason. As shown in Fig.~\ref{fig:supp-drscore}, a higher score indicates better task completion, while a lower score indicates that the degradation or restoration artifacts remain. In this way, DR-Score provides a simple auxiliary signal for degradation removal evaluation.

\noindent\textbf{Stability of DR-Score.}
Because VLM outputs can be stochastic, we evaluate each image five times and report the average score in all experiments. To verify stability, we calculate two complementary statistics.

(i) Run-level standard deviation (std). For each method, we first average scores over all test images within each run and then compute the std across repeated runs. The resulting std value is about 0.10--0.43, indicating that the mean scores at method-level are highly reproducible.

(ii) Image-level std. For each image, we compute the std across its repeated scores and then average over images. The std value is about 5.3--6.4, showing that individual-image scores can vary. 

Despite per-image score fluctuations, the std of DR-Scores at method-level is below 0.5, confirming that the relative ranking between methods is stable. We therefore report mean DR-scores in the experiments.

\noindent\textbf{Human Alignment.}
We then conduct a human study to examine how well different metrics agree with human judgment in UIR tasks, including DR-Score, MUSIQ \cite{musiq}, AFINE-NR \cite{afine}, TOPIQ \cite{chen2024topiq}, and MANIQA \cite{maniqa}.

The study is based on pairwise comparisons. For each test case, annotators are shown one LQ image and two restored results produced by different methods for the same task. They are asked to select the better result of the restoration task. The annotators are instructed to focus on three aspects: whether the target degradation is removed, whether the main scene content is preserved, and whether obvious artifacts are suppressed.
We sample 20 cases for each degradation type from the real-world test sets, including deblur, dehaze, de-rainstreak, de-raindrop, desnow, and low-light enhancement. For each pair, two methods are randomly selected from retrained models (DA-CLIP, FoundIR, FoundIR-v2, FAPE-IR) and PixRestore. The two results are presented in random order to avoid position bias. We also hide the method names and do not show any metric values.

A total of 20 annotators participated in this study. After collecting the annotations, we average the human choices and convert them into pairwise preferences. We then compare these preferences with the pairwise ranking induced by each metric. A pair is counted as aligned if the image preferred by humans also receives a better metric score. We report the average alignment ratio in the left part of Fig.~\ref{fig:human-align}. DR-Score achieves the highest agreement (90.7\%) with human preference, showing that it better reflects task completion and degradation removal than existing no-reference quality metrics. This supports its use as a practical auxiliary metric for real-world UIR evaluation.

\noindent\textbf{Failure Cases.}
DR-Score still has several limitations. It can be less reliable when two restored results are visually very similar and differ only in subtle appearance factors, such as brightness, local contrast, or color tone, as shown in the right part of Fig.~\ref{fig:human-align}. In such cases, even human annotators need to inspect carefully, and the VLM may fail to tell the difference.

\begin{table*}[t]
  \caption{Detailed quantitative comparison on Deblur benchmarks. The best and second-best results are highlighted in {\textcolor{red}{\textbf{red}}} and {\textcolor{blue}{\textit{blue italic}}}, respectively. Methods marked with $^*$ are retrained under the same training setting as ours.}
  \label{tab:app_deblur}
  \centering
  \scriptsize
  \setlength{\tabcolsep}{2.0pt}
  \renewcommand{\arraystretch}{0.95}
  \setlength{\aboverulesep}{0.3pt}
  \setlength{\belowrulesep}{0.3pt}
  \resizebox{\textwidth}{!}{%
  \begin{tabular}{l*{12}{c}}
    \toprule
    Method & \multicolumn{6}{c}{GoPro} & \multicolumn{6}{c}{UHD-blur} \\
    \cmidrule(lr){2-7} \cmidrule(lr){8-13}
     & PSNR$\uparrow$ & SSIM$\uparrow$ & LPIPS$\downarrow$ & DISTS$\downarrow$ & MUSIQ$\uparrow$ & AFINE-NR$\downarrow$ & PSNR$\uparrow$ & SSIM$\uparrow$ & LPIPS$\downarrow$ & DISTS$\downarrow$ & MUSIQ$\uparrow$ & AFINE-NR$\downarrow$ \\
    \midrule
    PromptIR & 23.43 & 0.7903 & 0.3035 & 0.1972 & 25.50 & -0.5874 & 24.21 & 0.7265 & 0.3228 & 0.2418 & 29.81 & -0.6163 \\
    PromptIR$^*$ & 29.82 & 0.8775 & 0.2000 & 0.1440 & 36.18 & -0.7123 & 28.39 & 0.8172 & 0.2101 & 0.1681 & 40.52 & -0.7570 \\
    DiffUIR & 29.32 & 0.8669 & 0.2041 & 0.1492 & 35.93 & -0.7137 & 26.39 & 0.7731 & 0.2476 & 0.1907 & 38.01 & -0.7075 \\
    UniRestore & 24.11 & 0.7496 & 0.2216 & 0.1408 & 45.93 & -0.7896 & 23.63 & 0.6914 & 0.2595 & 0.2024 & 42.99 & -0.6870 \\
    DA-CLIP & 28.57 & 0.8554 & 0.1279 & 0.0994 & 41.05 & -0.7329 & 26.23 & 0.7671 & 0.2003 & 0.1612 & 42.83 & -0.7479 \\
    DA-CLIP$^*$ & 28.87 & 0.8536 & 0.1317 & 0.1031 & 42.91 & -0.7444 & 27.71 & 0.7920 & 0.1630 & 0.1264 & 47.05 & -0.7656 \\
    FoundIR & 27.02 & 0.8121 & 0.2610 & 0.1800 & 29.96 & -0.6276 & 27.56 & 0.7971 & 0.2250 & 0.1762 & 38.80 & -0.7544 \\
    FoundIR$^*$ & 29.95 & 0.8762 & 0.1887 & 0.1393 & 35.81 & -0.7297 & 28.33 & 0.8170 & 0.2084 & 0.1613 & 39.98 & -0.7696 \\
    FoundIR-v2 & 24.48 & 0.7179 & 0.2224 & 0.1451 & \textcolor{blue}{\textit{57.73}} & \textcolor{blue}{\textit{-0.8954}} & 24.32 & 0.6980 & 0.2183 & 0.1680 & \textcolor{red}{\textbf{63.23}} & \textcolor{red}{\textbf{-1.0422}} \\
    FoundIR-v2$^*$ & 24.98 & 0.7470 & 0.1965 & 0.1262 & 56.90 & -0.8914 & 24.97 & 0.7232 & 0.1862 & 0.1385 & \textcolor{blue}{\textit{58.39}} & \textcolor{blue}{\textit{-0.9774}} \\
    Flux-IR & 24.03 & 0.7012 & 0.2408 & 0.1550 & 55.90 & -0.8500 & 22.98 & 0.6479 & 0.3084 & 0.2153 & 53.29 & -0.8064 \\
    Flux-IR$^*$ & 22.59 & 0.6716 & 0.2717 & 0.1936 & \textcolor{red}{\textbf{60.22}} & \textcolor{red}{\textbf{-0.9609}} & 21.71 & 0.6176 & 0.2918 & 0.2137 & 58.22 & -0.9313 \\
    FAPE-IR & 28.02 & 0.8374 & 0.1527 & 0.1058 & 41.62 & -0.7781 & 25.58 & 0.7417 & 0.2669 & 0.2054 & 35.00 & -0.7267 \\
    FAPE-IR$^*$ & 28.68 & 0.8505 & 0.1347 & 0.0907 & 38.84 & -0.7403 & 27.25 & 0.7908 & 0.1807 & 0.1282 & 41.57 & -0.7694 \\
    \rowcolor{lotuspink} \textbf{PixRestore-S} & 28.96 & 0.8553 & 0.1067 & 0.0833 & 43.39 & -0.8182 & 27.67 & 0.8016 & 0.1335 & 0.1046 & 49.73 & -0.8105 \\
    \rowcolor{lotuspink} \textbf{PixRestore-B} & 30.00 & 0.8805 & 0.0895 & 0.0721 & 44.74 & -0.8354 & 28.46 & 0.8218 & 0.1206 & 0.0949 & 49.54 & -0.8200 \\
    \rowcolor{lotuspink} \textbf{PixRestore-L} & \textcolor{blue}{\textit{30.79}} & \textcolor{blue}{\textit{0.8948}} & \textcolor{blue}{\textit{0.0787}} & \textcolor{blue}{\textit{0.0657}} & 45.51 & -0.8616 & \textcolor{blue}{\textit{28.96}} & \textcolor{blue}{\textit{0.8346}} & \textcolor{blue}{\textit{0.1117}} & \textcolor{blue}{\textit{0.0891}} & 50.45 & -0.8515 \\
    \rowcolor{lotuspink} \textbf{PixRestore-XL} & \textcolor{red}{\textbf{31.23}} & \textcolor{red}{\textbf{0.9025}} & \textcolor{red}{\textbf{0.0721}} & \textcolor{red}{\textbf{0.0621}} & 46.34 & -0.8712 & \textcolor{red}{\textbf{29.07}} & \textcolor{red}{\textbf{0.8407}} & \textcolor{red}{\textbf{0.1090}} & \textcolor{red}{\textbf{0.0854}} & 50.34 & -0.8487 \\
    \bottomrule
  \end{tabular}%
  }
\end{table*}

% ---- preamble helpers ----
% \usepackage{booktabs,xcolor,colortbl}
% \definecolor{lotuspink}{RGB}{255,230,240}
% -----------------------------

\begin{table*}[t]
  \caption{Detailed quantitative comparison on Dehaze benchmarks. The best and second-best results are highlighted in {\textcolor{red}{\textbf{red}}} and {\textcolor{blue}{\textit{blue italic}}}, respectively. Methods marked with $^*$ are retrained under the same training setting as ours.}
  \label{tab:app_dehaze}
  \centering
  \scriptsize
  \setlength{\tabcolsep}{2.0pt}
  \renewcommand{\arraystretch}{0.95}
  \setlength{\aboverulesep}{0.3pt}
  \setlength{\belowrulesep}{0.3pt}
  \resizebox{\textwidth}{!}{%
  \begin{tabular}{l*{12}{c}}
    \toprule
    Method & \multicolumn{6}{c}{RESIDE-6K} & \multicolumn{6}{c}{UHD-Haze} \\
    \cmidrule(lr){2-7} \cmidrule(lr){8-13}
     & PSNR$\uparrow$ & SSIM$\uparrow$ & LPIPS$\downarrow$ & DISTS$\downarrow$ & MUSIQ$\uparrow$ & AFINE-NR$\downarrow$ & PSNR$\uparrow$ & SSIM$\uparrow$ & LPIPS$\downarrow$ & DISTS$\downarrow$ & MUSIQ$\uparrow$ & AFINE-NR$\downarrow$ \\
    \midrule
    PromptIR & 26.69 & 0.9572 & 0.0460 & 0.0418 & 51.24 & -0.9422 & 15.98 & 0.8035 & 0.2391 & 0.1605 & 63.94 & -0.9765 \\
    PromptIR$^*$ & 24.01 & 0.9417 & 0.0642 & 0.0553 & 51.58 & -0.9362 & 18.49 & 0.8445 & 0.2054 & 0.1216 & 63.45 & -1.0058 \\
    DiffUIR & 24.66 & 0.9311 & 0.0708 & 0.0582 & 50.59 & -0.9314 & 16.15 & 0.8001 & 0.2572 & 0.1768 & 62.57 & -0.9683 \\
    UniRestore & 23.63 & 0.9162 & 0.1171 & 0.0869 & 55.68 & -0.9630 & 16.59 & 0.7776 & 0.3073 & 0.1849 & 62.14 & -0.9155 \\
    DA-CLIP & 28.66 & 0.9273 & 0.0559 & 0.0463 & 53.52 & \textcolor{blue}{\textit{-0.9784}} & 17.27 & 0.8230 & 0.2063 & 0.1307 & 65.88 & -1.0698 \\
    DA-CLIP$^*$ & 28.15 & 0.9579 & 0.0431 & 0.0409 & 51.31 & -0.9439 & 15.70 & 0.7979 & 0.2487 & 0.1746 & 63.98 & -0.9540 \\
    FoundIR & 16.52 & 0.8381 & 0.1665 & 0.1304 & 49.68 & -0.8612 & 13.62 & 0.7421 & 0.3499 & 0.2533 & 59.75 & -0.8267 \\
    FoundIR$^*$ & 25.07 & 0.9525 & 0.0523 & 0.0483 & 50.61 & -0.9464 & 16.91 & 0.8282 & 0.2146 & 0.1441 & 64.37 & -0.9976 \\
    FoundIR-v2 & 19.01 & 0.8101 & 0.1837 & 0.1318 & \textcolor{blue}{\textit{57.83}} & -0.9396 & 19.11 & 0.7309 & 0.2018 & 0.1357 & \textcolor{red}{\textbf{69.04}} & -1.0056 \\
    FoundIR-v2$^*$ & 18.57 & 0.7925 & 0.1997 & 0.1342 & 56.12 & -0.9191 & 19.62 & 0.7365 & 0.1870 & 0.1222 & \textcolor{blue}{\textit{68.08}} & -1.0543 \\
    Flux-IR & 15.64 & 0.7796 & 0.2591 & 0.1639 & \textcolor{red}{\textbf{65.46}} & \textcolor{red}{\textbf{-1.0185}} & 13.83 & 0.7402 & 0.3361 & 0.2333 & 61.74 & -0.8161 \\
    Flux-IR$^*$ & 17.27 & 0.8416 & 0.1578 & 0.1074 & 50.54 & -0.8348 & 14.65 & 0.7706 & 0.3052 & 0.2083 & 60.48 & -0.8251 \\
    FAPE-IR & \textcolor{blue}{\textit{31.36}} & 0.9628 & 0.0378 & \textcolor{blue}{\textit{0.0364}} & 50.35 & -0.9573 & 19.20 & 0.8386 & 0.1542 & 0.0937 & 66.12 & -1.1555 \\
    FAPE-IR$^*$ & 28.90 & 0.9575 & 0.0411 & 0.0385 & 50.32 & -0.9488 & 21.78 & 0.8537 & 0.1442 & 0.0905 & 65.45 & -1.1361 \\
    \rowcolor{lotuspink} \textbf{PixRestore-S} & 28.41 & 0.9555 & 0.0473 & 0.0444 & 52.81 & -0.9777 & 22.51 & 0.8729 & 0.1318 & 0.0855 & 66.98 & -1.1836 \\
    \rowcolor{lotuspink} \textbf{PixRestore-B} & 29.87 & 0.9636 & 0.0390 & 0.0388 & 51.79 & -0.9737 & \textcolor{red}{\textbf{23.41}} & 0.8820 & \textcolor{red}{\textbf{0.1188}} & \textcolor{red}{\textbf{0.0760}} & 67.26 & \textcolor{red}{\textbf{-1.2451}} \\
    \rowcolor{lotuspink} \textbf{PixRestore-L} & 30.63 & \textcolor{blue}{\textit{0.9660}} & \textcolor{blue}{\textit{0.0365}} & 0.0366 & 52.11 & -0.9772 & \textcolor{blue}{\textit{23.23}} & \textcolor{red}{\textbf{0.8836}} & 0.1227 & \textcolor{blue}{\textit{0.0771}} & 66.81 & \textcolor{blue}{\textit{-1.2381}} \\
    \rowcolor{lotuspink} \textbf{PixRestore-XL} & \textcolor{red}{\textbf{31.73}} & \textcolor{red}{\textbf{0.9677}} & \textcolor{red}{\textbf{0.0340}} & \textcolor{red}{\textbf{0.0355}} & 51.83 & -0.9741 & 23.14 & \textcolor{blue}{\textit{0.8827}} & \textcolor{blue}{\textit{0.1223}} & 0.0776 & 66.92 & -1.2237 \\
    \bottomrule
  \end{tabular}%
  }
\end{table*}

% ---- preamble helpers ----
% \usepackage{booktabs,xcolor,colortbl}
% \definecolor{lotuspink}{RGB}{255,230,240}
% -----------------------------

\begin{table*}[t]
  \caption{Detailed quantitative comparison on Denoise benchmarks. The best and second-best results are highlighted in {\textcolor{red}{\textbf{red}}} and {\textcolor{blue}{\textit{blue italic}}}, respectively. Methods marked with $^*$ are retrained under the same training setting as ours.}
  \label{tab:app_denoise}
  \centering
  \scriptsize
  \setlength{\tabcolsep}{2.0pt}
  \renewcommand{\arraystretch}{0.95}
  \setlength{\aboverulesep}{0.3pt}
  \setlength{\belowrulesep}{0.3pt}
  \resizebox{\textwidth}{!}{%
  \begin{tabular}{l*{12}{c}}
    \toprule
    Method & \multicolumn{6}{c}{DIV2K (Gaussian)} & \multicolumn{6}{c}{PolyU} \\
    \cmidrule(lr){2-7} \cmidrule(lr){8-13}
     & PSNR$\uparrow$ & SSIM$\uparrow$ & LPIPS$\downarrow$ & DISTS$\downarrow$ & MUSIQ$\uparrow$ & AFINE-NR$\downarrow$ & PSNR$\uparrow$ & SSIM$\uparrow$ & LPIPS$\downarrow$ & DISTS$\downarrow$ & MUSIQ$\uparrow$ & AFINE-NR$\downarrow$ \\
    \midrule
    PromptIR & \textcolor{red}{\textbf{34.57}} & \textcolor{red}{\textbf{0.9079}} & 0.1411 & 0.1384 & \textcolor{blue}{\textit{64.24}} & -0.9098 & 30.50 & 0.8978 & 0.3506 & 0.1825 & 30.88 & -0.6068 \\
    PromptIR$^*$ & 33.91 & 0.8975 & 0.1463 & 0.1388 & 61.00 & -0.8517 & 37.00 & 0.9780 & 0.0744 & 0.0742 & 32.70 & -0.6156 \\
    DiffUIR & 21.24 & 0.7540 & 0.3437 & 0.2523 & 55.21 & -0.7013 & 31.98 & 0.9221 & 0.2769 & 0.1588 & 32.67 & -0.6270 \\
    UniRestore & 30.38 & 0.8728 & 0.1659 & 0.1420 & \textcolor{red}{\textbf{64.25}} & -0.8868 & 33.11 & 0.9209 & 0.3012 & 0.1918 & 31.60 & -0.5629 \\
    DA-CLIP & 28.91 & 0.7532 & 0.2412 & 0.1764 & 57.33 & -0.8155 & 25.79 & 0.8957 & 0.2506 & 0.1765 & 33.00 & -0.6137 \\
    DA-CLIP$^*$ & 30.76 & 0.7704 & 0.2448 & 0.1570 & 58.12 & -0.7929 & 37.19 & 0.9753 & 0.0477 & 0.0761 & 33.67 & -0.5932 \\
    FoundIR & 27.14 & 0.6061 & 0.4920 & 0.2574 & 45.98 & -0.6097 & \textcolor{blue}{\textit{37.77}} & \textcolor{blue}{\textit{0.9789}} & 0.0668 & 0.0707 & 33.55 & -0.6303 \\
    FoundIR$^*$ & \textcolor{blue}{\textit{34.06}} & \textcolor{blue}{\textit{0.8992}} & 0.1572 & 0.1469 & 62.82 & -0.9074 & \textcolor{red}{\textbf{38.25}} & \textcolor{red}{\textbf{0.9837}} & 0.0529 & 0.1022 & 34.24 & -0.6195 \\
    FoundIR-v2 & 25.43 & 0.6544 & 0.2658 & 0.1825 & 62.76 & -0.9197 & 28.31 & 0.8418 & 0.2913 & 0.2066 & \textcolor{red}{\textbf{52.61}} & \textcolor{red}{\textbf{-0.8119}} \\
    FoundIR-v2$^*$ & 25.96 & 0.6584 & 0.2488 & 0.1837 & 62.60 & -0.9080 & 30.39 & 0.8410 & 0.3290 & 0.1943 & 41.44 & -0.6478 \\
    Flux-IR & 21.33 & 0.4780 & 0.5758 & 0.2660 & 50.37 & -0.7322 & 30.08 & 0.8903 & 0.2637 & 0.2073 & \textcolor{blue}{\textit{43.26}} & \textcolor{blue}{\textit{-0.7832}} \\
    Flux-IR$^*$ & 24.34 & 0.5837 & 0.4296 & 0.2386 & 49.22 & -0.7338 & 29.26 & 0.8984 & 0.3281 & 0.1716 & 32.85 & -0.6348 \\
    FAPE-IR & 31.09 & 0.8540 & 0.1147 & 0.1013 & 60.49 & -0.8739 & 34.89 & 0.9636 & 0.1329 & 0.1212 & 35.29 & -0.6526 \\
    FAPE-IR$^*$ & 31.32 & 0.8572 & 0.1080 & 0.0978 & 60.52 & -0.8830 & 37.11 & 0.9772 & 0.0401 & \textcolor{red}{\textbf{0.0522}} & 33.52 & -0.6135 \\
    \rowcolor{lotuspink} \textbf{PixRestore-S} & 33.04 & 0.8923 & 0.0783 & 0.0863 & 63.62 & -0.9410 & 36.71 & 0.9749 & 0.0465 & \textcolor{blue}{\textit{0.0607}} & 34.54 & -0.6139 \\
    \rowcolor{lotuspink} \textbf{PixRestore-B} & 33.40 & 0.8968 & \textcolor{red}{\textbf{0.0725}} & 0.0777 & 63.84 & -0.9591 & 35.84 & 0.9745 & 0.0402 & 0.0636 & 34.15 & -0.6030 \\
    \rowcolor{lotuspink} \textbf{PixRestore-L} & 33.35 & 0.8962 & 0.0730 & \textcolor{red}{\textbf{0.0768}} & 64.12 & \textcolor{red}{\textbf{-0.9671}} & 35.91 & 0.9736 & \textcolor{blue}{\textit{0.0385}} & 0.0709 & 34.39 & -0.6022 \\
    \rowcolor{lotuspink} \textbf{PixRestore-XL} & 33.46 & 0.8979 & \textcolor{blue}{\textit{0.0729}} & \textcolor{blue}{\textit{0.0774}} & 64.06 & \textcolor{blue}{\textit{-0.9648}} & 36.50 & 0.9745 & \textcolor{red}{\textbf{0.0341}} & 0.0753 & 33.95 & -0.5994 \\
    \bottomrule
  \end{tabular}%
  }
\end{table*}

% ---- preamble helpers ----
% \usepackage{booktabs,xcolor,colortbl}
% \definecolor{lotuspink}{RGB}{255,230,240}
% -----------------------------

\begin{table*}[t]
  \caption{Detailed quantitative comparison on De-rainstreak benchmarks. The best and second-best results are highlighted in {\textcolor{red}{\textbf{red}}} and {\textcolor{blue}{\textit{blue italic}}}, respectively. Methods marked with $^*$ are retrained under the same training setting as ours.}
  \label{tab:app_derainstreak}
  \centering
  \scriptsize
  \setlength{\tabcolsep}{2.0pt}
  \renewcommand{\arraystretch}{0.95}
  \setlength{\aboverulesep}{0.3pt}
  \setlength{\belowrulesep}{0.3pt}
  \resizebox{\textwidth}{!}{%
  \begin{tabular}{l*{12}{c}}
    \toprule
    Method & \multicolumn{6}{c}{RainDS-real} & \multicolumn{6}{c}{RealRain-1K} \\
    \cmidrule(lr){2-7} \cmidrule(lr){8-13}
     & PSNR$\uparrow$ & SSIM$\uparrow$ & LPIPS$\downarrow$ & DISTS$\downarrow$ & MUSIQ$\uparrow$ & AFINE-NR$\downarrow$ & PSNR$\uparrow$ & SSIM$\uparrow$ & LPIPS$\downarrow$ & DISTS$\downarrow$ & MUSIQ$\uparrow$ & AFINE-NR$\downarrow$ \\
    \midrule
    PromptIR & 25.15 & 0.7483 & 0.2047 & 0.1392 & 60.54 & -0.8602 & 23.85 & 0.7531 & 0.5011 & 0.3370 & 42.82 & -0.6698 \\
    PromptIR$^*$ & 26.62 & 0.7933 & 0.1977 & 0.1254 & 61.79 & -0.8963 & 30.23 & 0.8845 & 0.3341 & 0.2554 & 36.49 & -0.5757 \\
    DiffUIR & 26.11 & 0.7888 & 0.1885 & 0.1168 & 63.88 & -0.9157 & 22.91 & 0.7348 & 0.5084 & 0.3329 & 45.21 & \textcolor{red}{\textbf{-0.7440}} \\
    UniRestore & 23.47 & 0.7143 & 0.3220 & 0.1838 & 65.71 & -0.8797 & 21.52 & 0.7439 & 0.5226 & 0.3540 & 45.97 & -0.7192 \\
    DA-CLIP & 24.66 & 0.7467 & 0.1834 & 0.1213 & 63.20 & -0.9041 & 24.35 & 0.7654 & 0.4861 & 0.3139 & 46.39 & -0.7354 \\
    DA-CLIP$^*$ & 25.72 & 0.7516 & 0.1432 & 0.0859 & 63.31 & -0.9261 & 37.50 & 0.9695 & 0.0665 & \textcolor{blue}{\textit{0.0910}} & 32.27 & -0.6034 \\
    FoundIR & 26.78 & 0.7890 & 0.1630 & 0.1075 & 63.26 & -0.9045 & 26.97 & 0.8635 & 0.3278 & 0.2523 & 39.15 & -0.6486 \\
    FoundIR$^*$ & 27.27 & \textcolor{red}{\textbf{0.8120}} & 0.1993 & 0.1191 & \textcolor{red}{\textbf{67.29}} & -0.9812 & 37.44 & 0.9655 & 0.1193 & 0.1216 & 35.47 & -0.6340 \\
    FoundIR-v2 & 23.65 & 0.6078 & 0.1956 & 0.1111 & 63.58 & -0.9722 & 22.69 & 0.7226 & 0.5362 & 0.3326 & \textcolor{blue}{\textit{48.23}} & \textcolor{blue}{\textit{-0.7375}} \\
    FoundIR-v2$^*$ & 23.74 & 0.6043 & 0.1906 & 0.1088 & 64.82 & -1.0001 & 31.96 & 0.9153 & 0.1641 & 0.1556 & 36.17 & -0.6297 \\
    Flux-IR & 22.34 & 0.6665 & 0.2695 & 0.1769 & 64.41 & -0.9676 & 19.63 & 0.5801 & 0.6554 & 0.3952 & \textcolor{red}{\textbf{50.43}} & -0.7374 \\
    Flux-IR$^*$ & 21.66 & 0.6410 & 0.2912 & 0.1803 & 62.00 & -0.9306 & 20.37 & 0.6278 & 0.6110 & 0.3840 & 45.93 & -0.6811 \\
    FAPE-IR & 26.53 & 0.7636 & 0.1407 & 0.0835 & 62.03 & -0.9748 & 28.50 & 0.8817 & 0.3232 & 0.2523 & 42.51 & -0.7340 \\
    FAPE-IR$^*$ & 26.67 & 0.7642 & 0.1270 & 0.0746 & 61.14 & -0.9816 & 37.14 & 0.9748 & 0.0535 & \textcolor{red}{\textbf{0.0778}} & 31.15 & -0.6092 \\
    \rowcolor{lotuspink} \textbf{PixRestore-S} & 27.06 & 0.7950 & 0.1180 & 0.0773 & \textcolor{blue}{\textit{66.07}} & -1.0433 & 37.50 & 0.9744 & 0.0625 & 0.1038 & 32.07 & -0.6152 \\
    \rowcolor{lotuspink} \textbf{PixRestore-B} & 27.41 & 0.8038 & 0.1078 & 0.0693 & 65.65 & -1.0461 & 38.28 & \textcolor{blue}{\textit{0.9758}} & 0.0456 & 0.0941 & 32.27 & -0.6153 \\
    \rowcolor{lotuspink} \textbf{PixRestore-L} & \textcolor{blue}{\textit{27.54}} & 0.8065 & \textcolor{red}{\textbf{0.0990}} & \textcolor{blue}{\textit{0.0664}} & 65.60 & \textcolor{red}{\textbf{-1.0532}} & \textcolor{blue}{\textit{38.51}} & \textcolor{red}{\textbf{0.9760}} & \textcolor{blue}{\textit{0.0404}} & 0.0933 & 32.65 & -0.6170 \\
    \rowcolor{lotuspink} \textbf{PixRestore-XL} & \textcolor{red}{\textbf{27.58}} & \textcolor{blue}{\textit{0.8079}} & \textcolor{blue}{\textit{0.1000}} & \textcolor{red}{\textbf{0.0650}} & 65.18 & \textcolor{blue}{\textit{-1.0463}} & \textcolor{red}{\textbf{38.60}} & 0.9753 & \textcolor{red}{\textbf{0.0401}} & 0.1028 & 32.81 & -0.6160 \\
    \bottomrule
  \end{tabular}%
  }
\end{table*}

% ---- preamble helpers ----
% \usepackage{booktabs,xcolor,colortbl}
% \definecolor{lotuspink}{RGB}{255,230,240}
% -----------------------------

\begin{table*}[t]
  \caption{Detailed quantitative comparison on De-raindrop benchmarks. The best and second-best results are highlighted in {\textcolor{red}{\textbf{red}}} and {\textcolor{blue}{\textit{blue italic}}}, respectively. Methods marked with $^*$ are retrained under the same training setting as ours.}
  \label{tab:app_deraindrop}
  \centering
  \scriptsize
  \setlength{\tabcolsep}{2.0pt}
  \renewcommand{\arraystretch}{0.95}
  \setlength{\aboverulesep}{0.3pt}
  \setlength{\belowrulesep}{0.3pt}
  \resizebox{\textwidth}{!}{%
  \begin{tabular}{l*{12}{c}}
    \toprule
    Method & \multicolumn{6}{c}{RainDS-real} & \multicolumn{6}{c}{UAV-Rain1k} \\
    \cmidrule(lr){2-7} \cmidrule(lr){8-13}
     & PSNR$\uparrow$ & SSIM$\uparrow$ & LPIPS$\downarrow$ & DISTS$\downarrow$ & MUSIQ$\uparrow$ & AFINE-NR$\downarrow$ & PSNR$\uparrow$ & SSIM$\uparrow$ & LPIPS$\downarrow$ & DISTS$\downarrow$ & MUSIQ$\uparrow$ & AFINE-NR$\downarrow$ \\
    \midrule
    PromptIR & 20.70 & 0.7069 & 0.2773 & 0.1560 & 55.31 & -0.8088 & 16.92 & 0.6855 & 0.4007 & 0.2211 & 66.57 & -0.8022 \\
    PromptIR$^*$ & 24.53 & 0.7529 & 0.2637 & 0.1347 & 59.14 & -0.8560 & 22.84 & 0.8482 & 0.1835 & 0.1305 & 68.19 & -0.8475 \\
    DiffUIR & 20.56 & 0.6996 & 0.3119 & 0.1678 & 56.28 & -0.8034 & 17.13 & 0.7011 & 0.3790 & 0.2132 & 66.67 & -0.7942 \\
    UniRestore & 20.25 & 0.6906 & 0.3482 & 0.1888 & 58.44 & -0.8111 & 16.75 & 0.5847 & 0.4746 & 0.2683 & 65.96 & -0.7044 \\
    DA-CLIP & 22.99 & 0.7018 & 0.1763 & 0.0977 & 61.66 & -0.8832 & 17.26 & 0.6958 & 0.3667 & 0.2047 & 67.27 & -0.8080 \\
    DA-CLIP$^*$ & 24.14 & 0.7077 & 0.1570 & 0.0905 & 60.92 & -0.8855 & 23.01 & \textcolor{blue}{\textit{0.8755}} & 0.1079 & 0.0799 & 69.43 & -0.9228 \\
    FoundIR & 20.67 & 0.7171 & 0.3145 & 0.1749 & 59.34 & -0.8041 & 17.07 & 0.6773 & 0.4026 & 0.2285 & 67.57 & -0.7823 \\
    FoundIR$^*$ & 25.41 & \textcolor{red}{\textbf{0.7708}} & 0.2487 & 0.1319 & \textcolor{blue}{\textit{65.20}} & -0.9316 & 23.40 & 0.8729 & 0.1397 & 0.0973 & 69.58 & -0.9111 \\
    FoundIR-v2 & 20.22 & 0.5725 & 0.3079 & 0.1566 & 59.84 & -0.8704 & 19.03 & 0.5194 & 0.2505 & 0.1560 & 69.47 & -0.8975 \\
    FoundIR-v2$^*$ & 21.72 & 0.5641 & 0.2435 & 0.1255 & 64.75 & -0.9876 & 19.91 & 0.5280 & 0.2165 & 0.1332 & 69.30 & -0.9315 \\
    Flux-IR & 21.01 & 0.6540 & 0.2335 & 0.1221 & 63.08 & \textcolor{blue}{\textit{-1.0046}} & 16.87 & 0.6377 & 0.4113 & 0.2328 & 66.83 & -0.8472 \\
    Flux-IR$^*$ & 18.85 & 0.5672 & 0.3313 & 0.1964 & \textcolor{red}{\textbf{68.66}} & \textcolor{red}{\textbf{-1.1729}} & 16.76 & 0.5897 & 0.3218 & 0.2186 & \textcolor{red}{\textbf{72.29}} & \textcolor{red}{\textbf{-1.0842}} \\
    FAPE-IR & 24.71 & 0.7255 & 0.1841 & 0.0930 & 59.93 & -0.9452 & 18.06 & 0.6389 & 0.2598 & 0.1702 & 65.88 & -0.8070 \\
    FAPE-IR$^*$ & 25.27 & 0.7263 & 0.1502 & 0.0790 & 58.40 & -0.9358 & 22.49 & 0.7183 & 0.1607 & 0.1145 & 68.35 & -0.9177 \\
    \rowcolor{lotuspink} \textbf{PixRestore-S} & 25.42 & 0.7394 & 0.1326 & 0.0761 & 61.54 & -0.9911 & 23.54 & 0.8117 & 0.1190 & 0.1002 & 70.15 & -0.9816 \\
    \rowcolor{lotuspink} \textbf{PixRestore-B} & 25.76 & 0.7461 & 0.1245 & 0.0720 & 61.41 & -0.9928 & 24.65 & 0.8490 & 0.0928 & 0.0806 & 70.35 & -1.0176 \\
    \rowcolor{lotuspink} \textbf{PixRestore-L} & \textcolor{blue}{\textit{25.89}} & 0.7504 & \textcolor{blue}{\textit{0.1195}} & \textcolor{blue}{\textit{0.0686}} & 61.30 & -0.9963 & \textcolor{blue}{\textit{25.16}} & 0.8631 & \textcolor{blue}{\textit{0.0810}} & \textcolor{blue}{\textit{0.0719}} & \textcolor{blue}{\textit{70.63}} & -1.0379 \\
    \rowcolor{lotuspink} \textbf{PixRestore-XL} & \textcolor{red}{\textbf{26.01}} & \textcolor{blue}{\textit{0.7567}} & \textcolor{red}{\textbf{0.1171}} & \textcolor{red}{\textbf{0.0677}} & 61.36 & -0.9939 & \textcolor{red}{\textbf{25.69}} & \textcolor{red}{\textbf{0.8774}} & \textcolor{red}{\textbf{0.0715}} & \textcolor{red}{\textbf{0.0655}} & 70.55 & \textcolor{blue}{\textit{-1.0392}} \\
    \bottomrule
  \end{tabular}%
  }
\end{table*}

% ---- preamble helpers ----
% \usepackage{booktabs,xcolor,colortbl}
% \definecolor{lotuspink}{RGB}{255,230,240}
% -----------------------------

\begin{table*}[t]
  \caption{Detailed quantitative comparison on Low-light Enhancement benchmarks. The best and second-best results are highlighted in {\textcolor{red}{\textbf{red}}} and {\textcolor{blue}{\textit{blue italic}}}, respectively. Methods marked with $^*$ are retrained under the same training setting as ours.}
  \label{tab:app_lowlight}
  \centering
  \scriptsize
  \setlength{\tabcolsep}{2.0pt}
  \renewcommand{\arraystretch}{0.95}
  \setlength{\aboverulesep}{0.3pt}
  \setlength{\belowrulesep}{0.3pt}
  \resizebox{\textwidth}{!}{%
  \begin{tabular}{l*{12}{c}}
    \toprule
    Method & \multicolumn{6}{c}{UHD-LL} & \multicolumn{6}{c}{LOL} \\
    \cmidrule(lr){2-7} \cmidrule(lr){8-13}
     & PSNR$\uparrow$ & SSIM$\uparrow$ & LPIPS$\downarrow$ & DISTS$\downarrow$ & MUSIQ$\uparrow$ & AFINE-NR$\downarrow$ & PSNR$\uparrow$ & SSIM$\uparrow$ & LPIPS$\downarrow$ & DISTS$\downarrow$ & MUSIQ$\uparrow$ & AFINE-NR$\downarrow$ \\
    \midrule
    PromptIR & 11.82 & 0.5660 & 0.5088 & 0.3116 & 35.06 & -0.6577 & 9.17 & 0.3902 & 0.5732 & 0.4415 & 38.49 & -0.7859 \\
    PromptIR$^*$ & 25.33 & 0.8846 & 0.2303 & 0.1757 & 45.85 & -0.6794 & 10.50 & 0.4916 & 0.4824 & 0.3268 & 44.53 & -0.8607 \\
    DiffUIR & 17.67 & 0.5064 & 0.6057 & 0.3320 & 37.19 & -0.4779 & 25.76 & \textcolor{blue}{\textit{0.9099}} & 0.1581 & 0.1264 & 68.79 & -0.9938 \\
    UniRestore & 12.41 & 0.6102 & 0.4665 & 0.2903 & 36.98 & -0.6417 & 9.48 & 0.4274 & 0.5347 & 0.3426 & 43.85 & -0.7569 \\
    DA-CLIP & 20.51 & 0.7436 & 0.3675 & 0.2201 & 48.51 & -0.6172 & 23.99 & 0.8395 & 0.1263 & 0.1043 & \textcolor{red}{\textbf{74.18}} & -1.0050 \\
    DA-CLIP$^*$ & 16.64 & 0.7776 & 0.2632 & 0.2016 & 47.98 & -0.7589 & 19.38 & 0.8594 & 0.1626 & 0.1212 & 66.85 & -0.9450 \\
    FoundIR & 14.22 & 0.6984 & 0.3548 & 0.2386 & 42.38 & -0.7037 & 16.47 & 0.7963 & 0.2519 & 0.1877 & 65.83 & -1.0337 \\
    FoundIR$^*$ & 24.28 & 0.8885 & 0.2113 & 0.1687 & 51.48 & -0.8552 & 22.40 & \textcolor{red}{\textbf{0.9168}} & 0.1471 & 0.1191 & 71.35 & -1.0455 \\
    FoundIR-v2 & 16.40 & 0.7120 & 0.3643 & 0.2381 & \textcolor{red}{\textbf{60.71}} & \textcolor{red}{\textbf{-0.9313}} & 17.95 & 0.7776 & 0.2620 & 0.1674 & 67.72 & -1.0373 \\
    FoundIR-v2$^*$ & 17.72 & 0.7272 & 0.3541 & 0.2273 & \textcolor{blue}{\textit{55.85}} & \textcolor{blue}{\textit{-0.8751}} & 16.36 & 0.7341 & 0.2834 & 0.1929 & 62.16 & -0.9620 \\
    Flux-IR & 15.35 & 0.5519 & 0.5455 & 0.2965 & 38.47 & -0.5140 & 22.36 & 0.8525 & 0.1658 & 0.1032 & \textcolor{blue}{\textit{72.10}} & \textcolor{blue}{\textit{-1.1667}} \\
    Flux-IR$^*$ & 14.33 & 0.6028 & 0.4866 & 0.2852 & 37.25 & -0.5275 & 22.47 & 0.8369 & 0.1906 & 0.1167 & 70.91 & \textcolor{red}{\textbf{-1.2086}} \\
    FAPE-IR & 12.57 & 0.6315 & 0.3899 & 0.2669 & 39.07 & -0.7489 & \textcolor{red}{\textbf{26.34}} & 0.8942 & \textcolor{blue}{\textit{0.1261}} & 0.1006 & 66.74 & -1.0265 \\
    FAPE-IR$^*$ & \textcolor{blue}{\textit{26.55}} & \textcolor{red}{\textbf{0.8980}} & 0.1445 & 0.1096 & 51.54 & -0.8156 & 25.53 & 0.9049 & 0.1296 & 0.1040 & 65.50 & -1.0535 \\
    \rowcolor{lotuspink} \textbf{PixRestore-S} & 26.28 & 0.8889 & 0.1420 & 0.1060 & 53.31 & -0.8461 & 24.96 & 0.8899 & 0.1300 & 0.0957 & 66.07 & -1.0386 \\
    \rowcolor{lotuspink} \textbf{PixRestore-B} & 26.36 & 0.8886 & 0.1384 & 0.1010 & 53.45 & -0.8534 & 25.08 & 0.8983 & \textcolor{red}{\textbf{0.1212}} & \textcolor{red}{\textbf{0.0899}} & 65.88 & -1.0329 \\
    \rowcolor{lotuspink} \textbf{PixRestore-L} & \textcolor{red}{\textbf{26.89}} & 0.8928 & \textcolor{blue}{\textit{0.1326}} & \textcolor{red}{\textbf{0.0979}} & 54.53 & -0.8613 & 25.30 & 0.8960 & 0.1273 & \textcolor{blue}{\textit{0.0932}} & 64.39 & -1.0055 \\
    \rowcolor{lotuspink} \textbf{PixRestore-XL} & 26.37 & \textcolor{blue}{\textit{0.8934}} & \textcolor{red}{\textbf{0.1324}} & \textcolor{blue}{\textit{0.0982}} & 54.62 & -0.8686 & \textcolor{blue}{\textit{26.20}} & 0.8955 & 0.1286 & 0.0949 & 63.64 & -1.0015 \\
    \bottomrule
  \end{tabular}%
  }
\end{table*}

% ---- preamble helpers ----
% \usepackage{booktabs,xcolor,colortbl}
% \definecolor{lotuspink}{RGB}{255,230,240}
% -----------------------------

\begin{table*}[t]
  \caption{Detailed quantitative comparison on Desnow benchmarks. The best and second-best results are highlighted in {\textcolor{red}{\textbf{red}}} and {\textcolor{blue}{\textit{blue italic}}}, respectively. Methods marked with $^*$ are retrained under the same training setting as ours.}
  \label{tab:app_desnow}
  \centering
  \small
  \renewcommand{\arraystretch}{0.95}
  \begin{tabular}{l*{6}{c}}
    \toprule
    Method & \multicolumn{6}{c}{WeatherBench} \\
    \cmidrule(lr){2-7}
     & PSNR$\uparrow$ & SSIM$\uparrow$ & LPIPS$\downarrow$ & DISTS$\downarrow$ & MUSIQ$\uparrow$ & AFINE-NR$\downarrow$ \\
    \midrule
    PromptIR & 22.26 & 0.7939 & 0.2452 & 0.1672 & 45.60 & -0.6023 \\
    PromptIR$^*$ & 29.32 & 0.8532 & 0.1818 & 0.1402 & 46.30 & -0.6161 \\
    DiffUIR & 22.95 & 0.7948 & 0.2392 & 0.1667 & 48.10 & -0.6185 \\
    UniRestore & 22.33 & 0.7863 & 0.2464 & 0.1771 & 49.39 & -0.6451 \\
    DA-CLIP & 23.60 & 0.7971 & 0.2221 & 0.1558 & 46.48 & -0.6127 \\
    DA-CLIP$^*$ & 28.31 & 0.8282 & 0.1360 & 0.1021 & 48.82 & -0.6424 \\
    FoundIR & 23.03 & 0.7999 & 0.2406 & 0.1630 & 45.77 & -0.6029 \\
    FoundIR$^*$ & 29.82 & 0.8678 & 0.1524 & 0.1224 & 47.98 & -0.6908 \\
    FoundIR-v2 & 24.72 & 0.7347 & 0.2513 & 0.1671 & \textcolor{red}{\textbf{59.98}} & \textcolor{red}{\textbf{-0.8032}} \\
    FoundIR-v2$^*$ & 26.28 & 0.7690 & 0.1824 & 0.1311 & 54.44 & -0.7440 \\
    Flux-IR & 21.74 & 0.7231 & 0.3434 & 0.2204 & 56.08 & -0.6879 \\
    Flux-IR$^*$ & 21.79 & 0.6831 & 0.3023 & 0.1959 & \textcolor{blue}{\textit{57.03}} & \textcolor{blue}{\textit{-0.7504}} \\
    FAPE-IR & 26.02 & 0.8191 & 0.1759 & 0.1189 & 46.29 & -0.6280 \\
    FAPE-IR$^*$ & 30.19 & 0.8676 & 0.1136 & 0.0849 & 47.84 & -0.6595 \\
    \rowcolor{lotuspink} \textbf{PixRestore-S} & 31.26 & 0.8859 & 0.0853 & 0.0669 & 49.86 & -0.6829 \\
    \rowcolor{lotuspink} \textbf{PixRestore-B} & 31.93 & 0.8959 & 0.0688 & 0.0584 & 50.10 & -0.6825 \\
    \rowcolor{lotuspink} \textbf{PixRestore-L} & \textcolor{blue}{\textit{32.35}} & \textcolor{blue}{\textit{0.9039}} & \textcolor{blue}{\textit{0.0656}} & \textcolor{blue}{\textit{0.0569}} & 50.36 & -0.6859 \\
    \rowcolor{lotuspink} \textbf{PixRestore-XL} & \textcolor{red}{\textbf{32.57}} & \textcolor{red}{\textbf{0.9077}} & \textcolor{red}{\textbf{0.0623}} & \textcolor{red}{\textbf{0.0553}} & 50.25 & -0.6864 \\
    \bottomrule
  \end{tabular}%
  
\end{table*}

% ---- preamble helpers ----
% \usepackage{booktabs,xcolor,colortbl}
% \definecolor{lotuspink}{RGB}{255,230,240}
% -----------------------------

\begin{table*}[t]
  \caption{Detailed quantitative comparison on Super-resolution benchmarks. The best and second-best results are highlighted in {\textcolor{red}{\textbf{red}}} and {\textcolor{blue}{\textit{blue italic}}}, respectively. Methods marked with $^*$ are retrained under the same training setting as ours.}
  \label{tab:app_sr}
  \centering
  \scriptsize
  \setlength{\tabcolsep}{2.0pt}
  \renewcommand{\arraystretch}{0.95}
  \setlength{\aboverulesep}{0.3pt}
  \setlength{\belowrulesep}{0.3pt}
  \resizebox{\textwidth}{!}{%
  \begin{tabular}{l*{12}{c}}
    \toprule
    Method & \multicolumn{6}{c}{RealSR} & \multicolumn{6}{c}{ScreenSR} \\
    \cmidrule(lr){2-7} \cmidrule(lr){8-13}
     & PSNR$\uparrow$ & SSIM$\uparrow$ & LPIPS$\downarrow$ & DISTS$\downarrow$ & MUSIQ$\uparrow$ & AFINE-NR$\downarrow$ & PSNR$\uparrow$ & SSIM$\uparrow$ & LPIPS$\downarrow$ & DISTS$\downarrow$ & MUSIQ$\uparrow$ & AFINE-NR$\downarrow$ \\
    \midrule
    PromptIR & 23.47 & 0.7380 & 0.4647 & 0.2670 & 25.95 & -0.4435 & 25.05 & 0.7365 & 0.4140 & 0.2351 & 42.95 & -0.7419 \\
    PromptIR$^*$ & 28.65 & \textcolor{blue}{\textit{0.8051}} & 0.3045 & 0.2382 & 45.99 & -0.6821 & \textcolor{blue}{\textit{26.54}} & \textcolor{blue}{\textit{0.7884}} & 0.2633 & 0.2033 & 58.41 & -0.8518 \\
    DiffUIR & 27.40 & 0.7734 & 0.3744 & 0.2417 & 35.15 & -0.4999 & 25.68 & 0.7431 & 0.3989 & 0.2338 & 48.34 & -0.7720 \\
    UniRestore & 24.84 & 0.7683 & 0.3358 & 0.2305 & 39.91 & -0.5857 & 24.76 & 0.7403 & 0.3737 & 0.2263 & 53.30 & -0.8091 \\
    DA-CLIP & 24.21 & 0.7258 & 0.3910 & 0.2469 & 30.60 & -0.4804 & 23.24 & 0.6321 & 0.3635 & 0.2265 & 47.89 & -0.7682 \\
    DA-CLIP$^*$ & 27.72 & 0.7777 & 0.2213 & 0.1821 & 50.30 & -0.7260 & 25.26 & 0.7393 & 0.2469 & 0.1705 & 64.42 & -0.9523 \\
    FoundIR & 26.13 & 0.7432 & 0.4474 & 0.2619 & 26.74 & -0.4396 & 25.58 & 0.7365 & 0.4086 & 0.2365 & 42.63 & -0.7437 \\
    FoundIR$^*$ & \textcolor{blue}{\textit{28.74}} & 0.8000 & 0.3354 & 0.2454 & 40.54 & -0.6443 & \textcolor{red}{\textbf{26.56}} & \textcolor{red}{\textbf{0.7903}} & 0.2412 & 0.2048 & 61.52 & -0.9192 \\
    FoundIR-v2 & 24.61 & 0.6682 & 0.3303 & 0.2248 & 68.03 & -1.0134 & 23.09 & 0.6640 & 0.2616 & 0.1700 & 65.93 & -0.9700 \\
    FoundIR-v2$^*$ & 24.83 & 0.6649 & 0.3206 & 0.2179 & 64.84 & -0.9785 & 22.74 & 0.6411 & 0.1781 & \textcolor{red}{\textbf{0.1185}} & \textcolor{red}{\textbf{72.46}} & -1.0921 \\
    Flux-IR & 23.42 & 0.6599 & 0.3548 & 0.2465 & \textcolor{red}{\textbf{69.74}} & \textcolor{red}{\textbf{-1.0971}} & 21.56 & 0.6483 & 0.2258 & 0.1664 & \textcolor{blue}{\textit{72.17}} & \textcolor{red}{\textbf{-1.2027}} \\
    Flux-IR$^*$ & 22.41 & 0.5927 & 0.3668 & 0.2559 & \textcolor{blue}{\textit{68.62}} & \textcolor{blue}{\textit{-1.0386}} & 19.34 & 0.5778 & 0.3471 & 0.2479 & 71.41 & \textcolor{blue}{\textit{-1.1189}} \\
    FAPE-IR & 27.92 & 0.7969 & 0.2325 & 0.1877 & 50.67 & -0.8189 & 25.17 & 0.7495 & 0.3307 & 0.2077 & 52.83 & -0.8243 \\
    FAPE-IR$^*$ & \textcolor{red}{\textbf{29.06}} & \textcolor{red}{\textbf{0.8139}} & 0.1843 & 0.1468 & 50.37 & -0.8224 & 25.90 & 0.7546 & 0.2061 & 0.1398 & 61.45 & -0.8721 \\
    \rowcolor{lotuspink} \textbf{PixRestore-S} & 28.40 & 0.7882 & 0.1776 & 0.1458 & 56.20 & -0.8501 & 25.62 & 0.7578 & 0.1696 & 0.1258 & 66.40 & -0.9682 \\
    \rowcolor{lotuspink} \textbf{PixRestore-B} & 28.43 & 0.7904 & 0.1646 & 0.1367 & 56.15 & -0.8590 & 25.71 & 0.7642 & 0.1556 & \textcolor{blue}{\textit{0.1211}} & 67.20 & -0.9940 \\
    \rowcolor{lotuspink} \textbf{PixRestore-L} & 28.55 & 0.7933 & \textcolor{red}{\textbf{0.1590}} & \textcolor{red}{\textbf{0.1317}} & 57.26 & -0.8880 & 25.46 & 0.7508 & \textcolor{red}{\textbf{0.1483}} & 0.1227 & 68.72 & -0.9846 \\
    \rowcolor{lotuspink} \textbf{PixRestore-XL} & 28.59 & 0.7956 & \textcolor{blue}{\textit{0.1595}} & \textcolor{blue}{\textit{0.1332}} & 56.37 & -0.8922 & 25.17 & 0.7399 & \textcolor{blue}{\textit{0.1504}} & 0.1353 & 68.07 & -0.9560 \\
    \bottomrule
  \end{tabular}%
  }
\end{table*}

\begin{figure*}[h]
  \centering
  \includegraphics[width=0.9\linewidth]{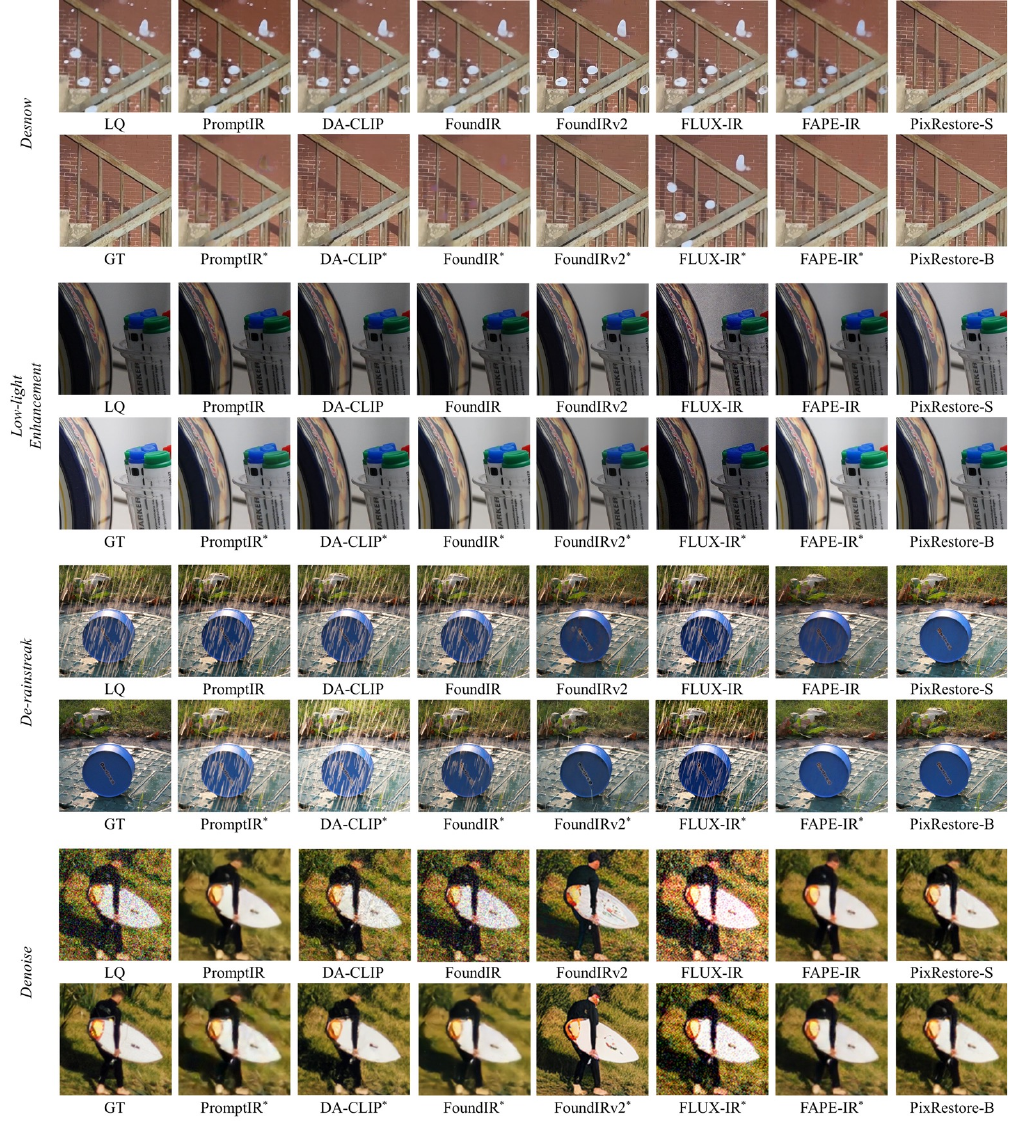}
  \vspace{-4mm}
  \caption{Visual comparisons on synthetic desnow, low-light enhancement, de-rainstreak, and denoise cases. Overall, PixRestore restores cleaner and more faithful results with better structural details and fewer artifacts.}
  \label{fig:supp-syn1}
  \vspace{-5mm}
\end{figure*}

\begin{figure*}[h]
  \centering
  \includegraphics[width=0.9\linewidth]{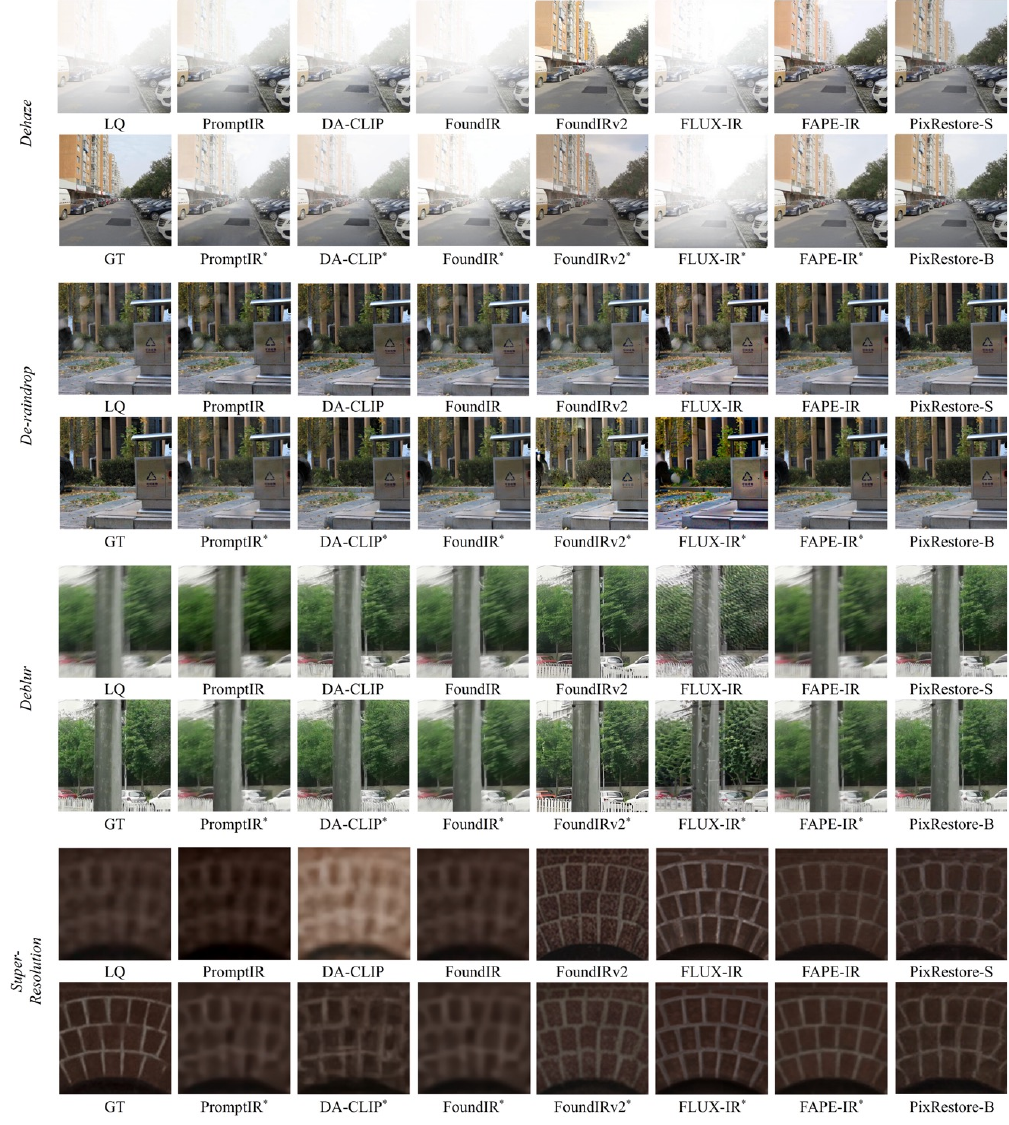}
  \vspace{-4mm}
  \caption{Visual comparisons on synthetic dehaze, de-raindrop, deblur, and SR cases. Overall, PixRestore restores cleaner and more faithful results with better structural details and fewer artifacts.}
  \label{fig:supp-syn2}
  \vspace{-5mm}
\end{figure*}

\section{More Public Benchmark Comparisons}

In the main paper, we report the average results of each degradation type. In this appendix, we further provide detailed comparisons on each benchmark, including GoPro \cite{GoPro} and UHD-Blur \cite{uhdblurhaze} for deblurring, RESIDE-6K \cite{RESIDE} and UHD-Haze \cite{uhdblurhaze} for dehazing, DIV2K \cite{div2k} with Gaussian noise and PolyU \cite{polyunoise} for denoising, RainDS-real \cite{rainds} and RealRain-1k \cite{realrain} for rain streak removal, RainDS-real \cite{rainds} and UAV-Rain1k \cite{UAV-Rain1k} for raindrop removal, UHD-LL \cite{uhdll} and LOL \cite{LoL} for low-light enhancement, WeatherBench \cite{guan2025weatherbench} for desnowing, and RealSR \cite{realsr} and ScreenSR \cite{vosr} for super-resolution. All images are center-cropped to 512 for testing.

The results are shown in Tables~\ref{tab:app_deblur}--\ref{tab:app_sr}. We see that PixRestore is not limited to a specific test dataset. It achieves strong and balanced performance across diverse restoration benchmarks. Some previous methods can obtain good no-reference scores by producing sharper or more contrastive outputs, but they often fall behind on fidelity-oriented full-reference metrics. For example, in the deblur task, FoundIR-v2 and Flux-IR$^*$ obtain much higher MUSIQ and better AFINE-NR on GoPro and UHD-Blur, but their PSNR, SSIM, LPIPS, and DISTS are clearly worse than PixRestore. Compared with previous UIR methods, our model consistently ranks among the top methods on distortion metrics such as PSNR, SSIM, LPIPS, and DISTS. At the same time, it remains competitive on no-reference metrics such as MUSIQ and AFINE-NR.

We can see that there is a clear trend across almost all tasks: scaling the model size is beneficial. From PixRestore-S to PixRestore-B, to PixRestore-L, and to PixRestore-XL, scaling generally improves average performance, although some individual datasets show non-monotonic behavior. This trend is especially clear for deblur, dehaze, desnow, and de-raindrop, where larger models repeatedly deliver stronger restoration fidelity. Although the gains on MUSIQ or AFINE-NR are sometimes less monotonic, the larger variants still show more stable top-tier performance overall. These results suggest that PixRestore scales well, and that increasing model capacity is an effective way to improve UIR performance.

The results of retrained models using our training data also reveal an important pattern. Many existing methods retrained on our training data improve their performance on almost all datasets. Nonetheless, these retrained baselines remain behind PixRestore on the main full-reference metrics. This suggests that using better training data alone is not sufficient; the model design itself also matters.

We provide more visual comparisons in Figs.~\ref{fig:supp-syn1} and \ref{fig:supp-syn2}. Overall, the compared methods show different trade-offs between degradation removal and detail preservation. PixRestore consistently produces cleaner and more balanced results across diverse synthetic tasks. For example, in the desnow case of Fig.~\ref{fig:supp-syn1}, PixRestore removes snow more thoroughly while preserving fine fence structures. Similar trends can be observed in Fig.~\ref{fig:supp-syn2}. In deblurring, PixRestore restores sharper pole boundaries and cleaner background tree textures; in super-resolution, it recovers clearer brick patterns and more faithful structures than competing methods.

\end{document}